\documentclass[10pt,journal,compsoc]{IEEEtran}
\usepackage{graphicx}
\usepackage{amsmath,amssymb} 
\usepackage{color}
\usepackage{todonotes}

\usepackage{algorithm}
\usepackage{algpseudocode}  
\usepackage{multirow}
\usepackage{booktabs}
\usepackage{subfigure}
\usepackage{hyperref}
\usepackage{color,soul}

\ifCLASSOPTIONcompsoc
  \usepackage[nocompress]{cite}
\else
  \usepackage{cite}
\fi

\ifCLASSINFOpdf
\else
\fi

\begin{document}

%
\title{Two-stage Temporal Modelling Framework for Video-based Depression Recognition using Graph Representation}
%
%
%
%
\author{Jiaqi Xu,~\IEEEmembership{}
        Siyang Song,~\IEEEmembership{}
        Keerthy Kusumam,~\IEEEmembership{}
        Hatice Gunes~\IEEEmembership{}
        and Michel Valstar~~\IEEEmembership{}

\IEEEcompsocitemizethanks{\IEEEcompsocthanksitem Jiaqi Xu is with the Sensor Information Lab, Department of Automation, University of Harbin University of Science and Technology, Harbin 150080, China. E-mail: 192050041@stu.hrbust.edu.cn)

\IEEEcompsocthanksitem Siyang Song and Hatice Gunes are with the AFAR Lab, Department of Computer Science and Technology, University of Cambridge, Cambridge, CB3 0FT, United Kingdom.
E-mail: ss2796@cam.ac.uk, Hatice.Gunes@cl.cam.ac.uk
(Corresponding Author: Siyang Song, E-mail: ss2796@cam.ac.uk)

\IEEEcompsocthanksitem Keerthy Kusumam and Michel Valstar are with the Computer Vision Lab, School of Computer Science, University of Nottingham, Nottingham, NG8 1BB, United Kingdom. E-mail: \{keerthy.kusumam, michel.valstar\} @nottingham.ac.uk}

\thanks{Manuscript received May 5, 2020}}

%
%

\markboth{IEEE TRANSACTIONS on AFFECTIVE COMPUTING}%
{Shell \MakeLowercase{\textit{et al.}}: Bare Advanced Demo of IEEEtran.cls for IEEE Computer Society Journals}

\IEEEtitleabstractindextext{%
\begin{abstract}


\noindent Video-based automatic depression analysis provides a fast, objective and repeatable self-assessment solution, which has been widely developed in recent years. While depression clues may be reflected by human facial behaviours of various temporal scales, most existing approaches either focused on modelling depression from short-term or video-level facial behaviours. In this sense, we propose a two-stage framework that models depression severity from multi-scale short-term and video-level facial behaviours. The short-term depressive behaviour modelling stage first deep learns depression-related facial behavioural features from multiple short temporal scales, where a Depression Feature Enhancement (DFE) module is proposed to enhance the depression-related clues for all temporal scales and remove non-depression noises. Then, the video-level depressive behaviour modelling stage proposes two novel graph encoding strategies, i.e., Sequential Graph Representation (SEG) and Spectral Graph Representation (SPG), to re-encode all short-term features of the target video into a video-level graph representation, summarizing depression-related multi-scale video-level temporal information. As a result, the produced graph representations predict depression severity using both short-term and long-term facial beahviour patterns. The experimental results on AVEC 2013 and AVEC 2014 datasets show that the proposed DFE module constantly enhanced the depression severity estimation performance for various CNN models while the SPG is superior than other video-level modelling methods. More importantly, the result achieved for the proposed two-stage framework shows its promising and solid performance compared to widely-used one-stage modelling approaches.


\end{abstract}

}

\maketitle

\IEEEdisplaynontitleabstractindextext

%
\IEEEpeerreviewmaketitle

\ifCLASSOPTIONcompsoc
\IEEEraisesectionheading{\section{Introduction}\label{sec:introduction}}
\else
\section{Introduction}
\label{sec:introduction}
\fi

\begin{figure*}
\centering
\includegraphics[width=16.6cm]{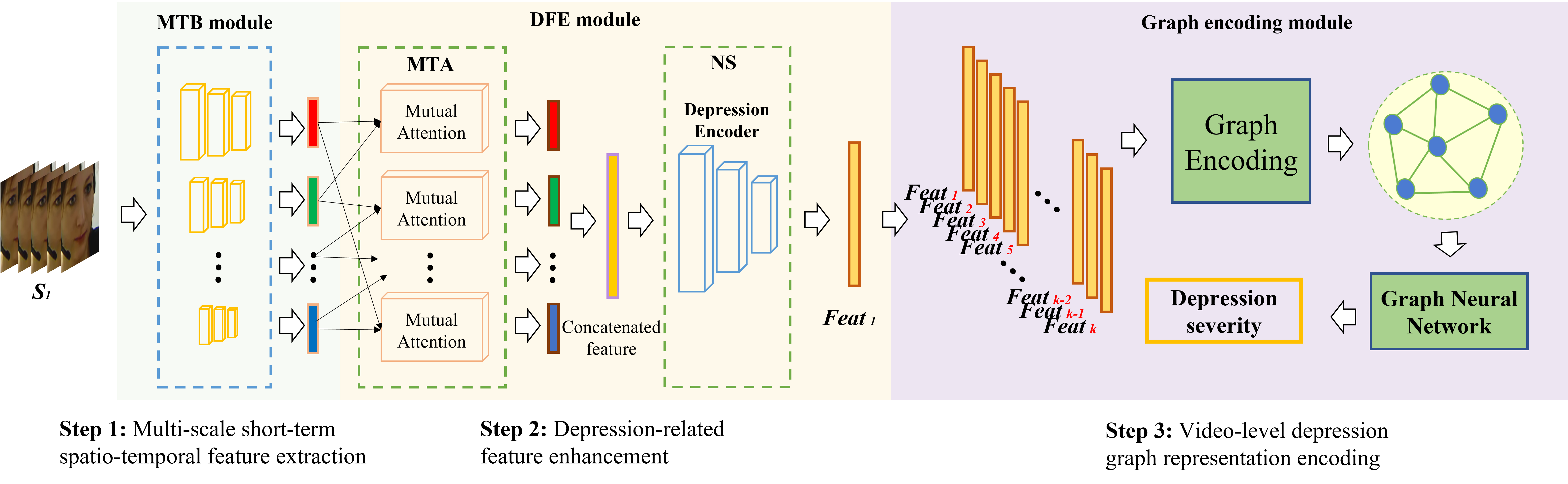}
\caption{The pipeline of the proposed approach which consists of three main modules. The MTB module first extracts short-term behavioural features at multiple spatio-temporal scales from every thin slice of the target video. Then, the DFE module enhances the depression-related cues encoded by the feature at each scale (MTA sub-module), respectively and disentangles non-depression noises in the concatenated feature (NS sub-module). Finally, we propose a graph encoding module to summarize short-term depression features learned from all thin slices of the target video into a video-level graph representation, and feed it to a Graph Neural Network (GNNs) for depression severity estimation.}
\label{fig:pipeline}
\end{figure*}

\noindent Major depressive disorder (MDD) is one of the most prevalent mental health issue that affects more than $2\%$ of the world population \cite{james2018global}, which is one of the major drivers that cause physical and mental disability, leading to severe consequences such as heart attacks and suicide \cite{american2013diagnostic}. While traditional clinical depression assessments require patients to fill in screening questionnaires or seek clinical support from a physician, such assessments are subjective and usually result in long waiting times causing delay in delivering treatment or intervention. Previous psychological studies have frequently shown that non-verbal facial behaviours are reliable markers of depression  \cite{chentsova2010further,ellgring2007non}. The recent advances in computer vision facilitate machines to automatically recognize human facial behaviours \cite{churamani2020clifer,li2020deep,song2021self,liu2020facial}, making it feasible to automatically analyze depression from face videos. As a result, face video-based automatic depression analysis has drawn considerable attention in the past decade \cite{valstar2013avec,ringeval2017avec,ringeval2019avec}.


Existing video-based automatic depression analysis approaches can be categorized into two groups: frame/thin slice-level modelling  methods and video-level modelling methods. The frame/thin slice-level modelling methods \cite{dibekliouglu2018dynamic,zhou2018visually,yang2018integrating,al2018video,uddin2020depression,he2021automatic} individually infer depression status for each frame or thin slice of the video, primarily focusing on the depression-related cues from subjects' facial appearance or the short-term facial dynamics. Most of these approaches either disregard the temporal information or only consider single-scale short-term facial dynamics exhibited within a pre-defined time-window. Since facial dynamics are a key component of facial behaviours and given that depression-related cues may be encoded by the facial behaviours at varying temporal scales, such methods would miss crucial information at the feature extraction stage. Moreover, as discussed in \cite{song2020spectral}, only using short-term facial behaviours to infer depression is not reliable as similar short-term facial behaviours may be exhibited by subjects with different depression severity levels. Although some of these approaches \cite{al2018video,uddin2020depression,haque2018measuring} employ RNNs/LSTMs to learn long-term dependencies from the learned frame/thin slice-level features, regressors that are trained by pairing a frame/thin slice with the video-level depression label cannot learn a good hypothesis. This is because such a training strategy may lead the regressors to focus on learning non-depression related facial attributes that are invariant for the subject in each video, e.g., identity, rather than depression-related facial actions.



Since depression is a long-term mental state lasting much longer than the duration of a regular video (i.e., usually less than an hour \cite{valstar2013avec,valstar2014avec,ringeval2017avec,ringeval2019avec,jaiswal2019virtual}), many recent studies propose to infer depression based on the features that are extracted from an entire video \cite{song2020spectral,de2021mdn,he2018automatic,song2018human,niu2020multimodal,jain2014depression}. Most of these approaches surpass the performance of frame/thin slice-level modelling methods. However, hand-crafted methods \cite{meng2013depression,he2018automatic,jain2014depression,jaiswal2019automatic} which are engineered to summarize the frame-level/thin slice-level features into a video-level descriptor, generally fail to learn depression-specific features that can be learned using deep learning methods. A standard approach to apply deep learning to video-level depression-related descriptors is to select a fixed number of key-frames from each video, and then feed them to the 3D CNNs to learn a video-level depression descriptor \cite{de2021mdn}. However, these approaches discard a large number of frames, ignoring local short-term facial dynamics that may contain crucial information for depression analysis.

In this paper, we hypothesize that both short-term and video-level (long-term) facial behaviours encode depression-related cues and the optimal temporal scales for such information are not well defined. Motivated by this, we propose a specific, two-stage framework for video-based automatic depression analysis. The first short-term depressive behaviour modelling stage learns multi-scale short-term facial behaviour features from each thin slice of the target video and is designed to further enhance the depression-related cues whilst suppressing non-depression related noise at varying temporal scales. During the second video-level depressive behaviour modelling stage, we propose to represent the depression-related features encoded by the entire video using a graph representation, thereby summarising all thin slice-level features of the target video into a unified descriptor. In particular, We propose two novel graph encoding strategies: sequential graph representation (SEG) and spectral graph representation (SPG). Importantly, both of these methods encode multi-scale long-term and short-term facial dynamics of the target video that are learned from all the available frames without forgoing any details. The resulting graph representations can then be processed by Graph Neural Networks for depression analysis. The pipeline of the proposed approach is illustrated in Fig. \ref{fig:pipeline}. The main contributions of this paper are summarized as follows:

\begin{itemize}

    \item This paper proposes a specific, two-stage deep-learning framework for video-based automatic depression analysis which provides high performance gains in comparison to existing single-stage methods that only model depression at either frame/thin slice-level or video-level. We demonstrate the effectiveness of the two stage framework in our experiments. This framework can easily be extended by replacing the proposed short-term or video-level modules with more advanced or preferred components. 

    \item We propose a novel short-term behaviour modelling module (MTB-DFE) which can enhance the depression-related behaviour cues and disentangle non-depression noises from the features learned from multiple spatio-temporal scales.

    \item We propose a novel graph-based video-level modelling approach that summarizes all short-term depression-related features of the target video into a unified and length-independent video-level graph representation which not only encodes multi-scale short-term and long-term spatio-temporal behavioural dynamics but also utilizes all available frames of the video. To the best of our knowledge, this is the first work that applies Graph Neural Network (GNNs) for face video-based automatic depression analysis.
    
\end{itemize}

\section{Related Work}

\noindent In this section, we first briefly present the evidence from psychology literature supporting the notion that signs of depression can be reflected in human facial behaviours (Sec. \ref{subsec:relationship}). We then review the recently proposed video-based automatic depression analysis approaches in Sec. \ref{subsec:video_depression}. We also list, in particular, the existing methods that represent the human face as a graph in Sec. \ref{subsec:re_GNN}.

\subsection{Relationship between depression and facial behaviours}
\label{subsec:relationship}

\noindent Previous studies have shown that depression is well associated with human facial behaviours. One key finding is that depression is usually accompanied by the reduced facial displays of positive emotions, which has been frequently validated across various studies \cite{chentsova2010further,gehricke2000reduced,renneberg2005facial,gaebel2004facial}. In addition, the individuals diagnosed with depression usually have less facial expressiveness \cite{gaebel2004facial,renneberg2005facial} and head movements \cite{fisch1983analyzing,joshi2013can}.  Ellgring et al. \cite{ellgring2007non} have summarized typical symptoms of depression in terms of facial expressions, indicating that depression is not only associated with sorrowful facial displays but also with \textit{``a total lack of facial expressions corresponding to the lack of affective experience"}. Meanwhile, some previous studies \cite{cohn2009detecting,girard2014nonverbal} have particularly investigated the relationship between depression and standard facial action units (AUs). The results show that individuals that have high depression severity presented fewer affiliative facial expressions (AU 12 and AU 15), but more non-affiliative facial expressions (AU 14) and diminished head motions.

\subsection{Video-based automatic depression analysis}
\label{subsec:video_depression}

\noindent Most existing video-based automatic depression analysis approaches are single-stage methods, i.e., extracting depression feature from a single frame/thin slice or the entire video. In particular, the frame/thin slice-level methods attempted to model depression status based on individuals' facial appearance \cite{zhou2018visually,uddin2020depression,he2021automatic} (e.g., frame-level modelling) or short-term facial behaviours (thin slice-level modelling) \cite{yang2018integrating,de2020deep,haque2018measuring,gupta2014multimodal,zhou2020facial,al2018video,zhu2017automated}. The frame-level modelling approaches usually focus on learning the depression-related salient facial appearance information. Zhou et al.\cite{zhou2018visually} identified the salient facial region for depression markers, where the depression-related facial regions of each frame are highlighted to predict depression. Meanwhile, the thin slice-level modelling approaches not only utilize facial appearance but also incorporate short-term facial dynamics. Such approaches usually divide each video into several equal-length segments, and learn depression features from each segment individually. A popular approach is to use a C3D network  \cite{al2018video,de2020deep} to extract spatio-temporal feature from thin video slices. For most frame-level and thin slice-level modelling approaches, the video-level prediction are aggregated by computing the average of all frame/slice predictions. As discussed in Sec. \ref{sec:introduction}, these methods fail to consider the important long-term facial behaviours/dynamics for depression recognition. Although some of the methods \cite{al2018video,uddin2020depression,al2018video,haque2018measuring} uses RNNs/LSTMs to model long-term temporal dependencies from the video, the CNNs of such methods are trained by pairing a frame/thin slice with the video-level label are problematic.

To avoid the ambiguity arising from frame/thin slice-level modelling approaches, many recent studies proposed to predict depression based on long-term behavioural information, e.g., learning a video-level depression-related feature. He et al. \cite{he2018automatic} extended the LBP-TOP feature to MRLBP-TOP for extracting short-term dynamics and then employs Fisher Vector to aggregate them as the long-term representation. Gong et al. \cite{gong2017topic} and Sun et al. \cite{sun2017random} investigated the relationship between the interview topics and depression severity level. Both methods built a topic-related descriptor for each video to infer depression severity. Besides the hand-crafted methods, De Melo et al. \cite{de2021mdn} proposed to down-sample the video into a small set of frames which roughly represent the video-level information and it was then fed to 3D CNNs to learn a video-level depression representation. Niu et al. \cite{niu2020multimodal} proposed a spatio-temporal attention network to integrate the facial appearance and short-term facial dynamics. Then, the eigen-evolution pooling strategy is introduced to aggregate thin slice-level features into the video-level descriptor. Song et al. \cite{song2018human,song2020spectral} represented a video as a low-dimensional multi-channel time-series signal and proposed a spectral approach to encode this time-series into a length-independent video-level spectral representation which contains multi-scale facial dynamics.

Although these deep learning-based approaches are capable of capturing video-level facial descriptors, they are also single-stage methods which falls short in specifically learning depression-related clues from short-term behavioural dynamics. While some of them \cite{al2018video,uddin2020depression,al2018video,haque2018measuring} using RNNs/LSTMs to model long-term temporal dependencies between the frame-level predictions of a video, they still have to pair each frame/thin slice with the video-level label during the training, resulting in the learned model to be problematic. Moreover, none of the above methods have investigated the idea of representing the video-level facial behaviours as a graph. In this paper, we propose a two-stage approach to model depression at both short-term and video-level, where video-level facial behaviours are encoded into a graph representation.






\subsection{Facial graph representation}
\label{subsec:re_GNN}

\noindent Many recent studies proposed to represent static facial appearance or spatio-temporal facial behaviours as a graph. The majority of static graph facial representations are either built on facial landmarks or facial regions. In such methods, the facial landmarks' coordinates \cite{zhang2020region,lei2020novel,hassan2020novel} or facial appearance features extracted from the facial regions \cite{liu2020relation,zhang2019context,xie2020adversarial} are used as the vertex features. The relationships between vertices are usually represented by an adjacency matrix, where a binary value (0 or 1) is employed to define the connectivity of each pair of vertices. In these methods, the adjacency matrix is obtained by the pre-computed relationships \cite{liu2020relation}, feature correlations/distances \cite{zhang2020region,fan2020facial} etc.

Few methods employ graph representations to learn spatio-temporal facial behaviours. In particular, some of the methods \cite{zhou2020facial,chen2019efficient} treat facial landmarks as the vertices, and extend the spatial facial graph to the spatio-temporal domain by constructing a spatial graph for each frame and then connecting them as a spatio-temporal graph, where the inter-frame edges connect the same node between consecutive frames. Another method \cite{liu2021video} constructs a facial sequential graph for each face sequence, where each frame is regarded as a vertex and the relationship between a pair of frames is defined as the corresponding edge feature.

However, none of the aforementioned approaches are suitable for constructing graph representations for long face videos, as the number of vertices and edges in spatio-temporal graph \cite{zhou2020facial,chen2019efficient} and the sequential graph \cite{liu2021video} would grow with the increasing number of the frames making them intractable for training. Motivated by this, in this paper, we propose the very first work, to the best of our knowledge, to construct a facial behavioural graph representation from a long video for automatic depression analysis.

\section{The proposed two-stage approach}

\noindent In this section, we present our two-stage framework, namely, the short-term depressive behaviour modelling stage and video-level depressive behaviour modelling stage. Our framework is designed to learn multi-scale short-term and long-term facial behaviour features for depression severity estimation, using all the available frames of the target video. The first stage (explained in Sec. \ref{subsec:TPAN}) of the proposed approach consists of two modules: (i). a Multi-scale Temporal Behavioural Feature Extraction Module (MTB) that learns short-term behavioural features at varying spatio-temporal scales, and (ii). a Depression Feature Enhancement (DFE) Module that enhances the depression-related cues and suppresses non-depression noises from the extracted behavioural features. Subsequently, for the video-level behaviour modelling stage (explained in Sec. \ref{subsec:graph_representation}) we propose two novel graph representations, each of which summarizes the extracted multi-scale short-term descriptors of the entire video into a video-level graph representation which encodes multi-scale depression-related cues. Finally, we feed the resulting graph representation to GNNs to provide a video-level depression prediction (Sec. \ref{subsec:graph_representation}).

The main contributions and benefits of our approach in comparison with the existing depression recognition approaches are the following: (i). In contrast to existing single-stage approaches that either focuses on modelling depression at frame/thin slice-level \cite{al2018video,zhou2018visually,yang2018integrating,uddin2020depression}  or video-level \cite{song2020spectral,de2021mdn,niu2020multimodal}, we propose a two-stage framework that takes advantage of both short-term and video-level behaviours for depression recognition; (ii) the framework is designed so that it utilizes all available frames to predict depression, distinguishing it from other video-level modelling methods \cite{de2021mdn} that discard frames carrying crucial information;  (iii). while widely-used C3D-based approaches \cite{de2021mdn,de2020deep,al2018video} only learn depression features based on a single temporal scale, the proposed short-term depressive behaviour modelling stage can explicitly encode depression-related facial behaviour features at multiple temporal scales; (iv). the proposed Depression Feature Enhancement (DFE) module is the very first work that is designed to specifically enhance the depression-related cues and suppress the non-depression noise for the deep-learned features; and (v). Compared to other video-level modelling methods \cite{song2020spectral,he2018automatic,jaiswal2019automatic,niu2020multimodal,jain2014depression} that simply employ statistics (e.g., the average value of frame-level predictions) to summarize the predictions/features of all frames/thin slices, we propose the first work that learns a graph representation to represent the video-level depression-related facial behaviours.


\begin{figure}
\centering
\includegraphics[width=8.6cm]{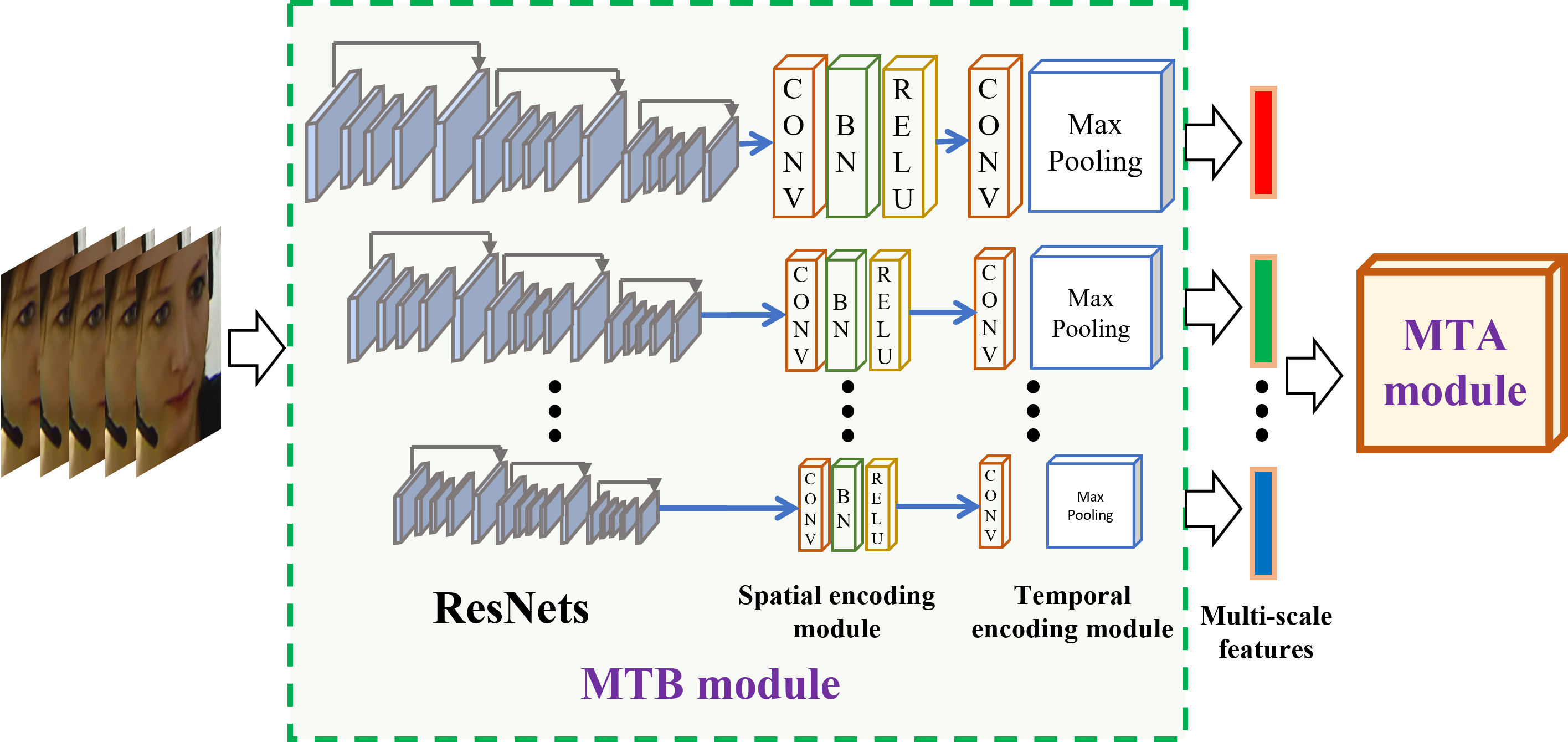}
\caption{The architecture of the Multi-scale Temporal Behavioural Feature Extraction (MTB) module.}
\label{fig:MTB}
\end{figure}

\subsection{Short-term depressive behaviour modelling}
\label{subsec:TPAN}


\noindent The following sections describe in detail the proposed short-term depressive behaviour modelling stage which consists of two modules: (i). a Multi-scale Temporal Behavioural Feature Extraction Module (MTB) and (ii). a Depression Feature Enhancement (DFE) Module.

\subsubsection{Multi-scale Temporal Behavioural Feature Extraction}

\noindent We build the MTB module based on the Temporal Pyramid Network (TPN) \cite{yang2020temporal}. As illustrated in Fig. \ref{fig:MTB}, the MTB consists of multiple branches that can learn multi-scale spatio-temporal features from an image sequence. Each branch consists of a single-depth 3D ResNet to produce feature maps from the input sequence at a unique spatio-temporal level. In particular, each branch first resizes the input sequence at a unique spatial scale and thus feature sequences of multiple spatial levels can be learned. Then, a spatial encoding module is attached to align spatial semantics of the produced feature sequences, each of which is then down-sampled by a pre-defined, unique temporal factors, respectively. In other words, a set of feature map sequences with temporal scales of $T_1, T_2, \cdots, T_K$ are produced. After that, a temporal encoding module is utilized to retrieve multi-scale depression-related behavioural temporal dynamics from the down-sampled feature sequences. As a result, the proposed MTB module can provide features that represent facial behaviours at multiple spatio-temporal scales for a thin video slice.

\subsubsection{Depression Feature Enhancement}

\begin{figure*}
	\centering
	\subfigure[The Mutual Temporal Attention (MTA) module]{\label{subfig:MTA}
		\includegraphics[width=16.5cm]{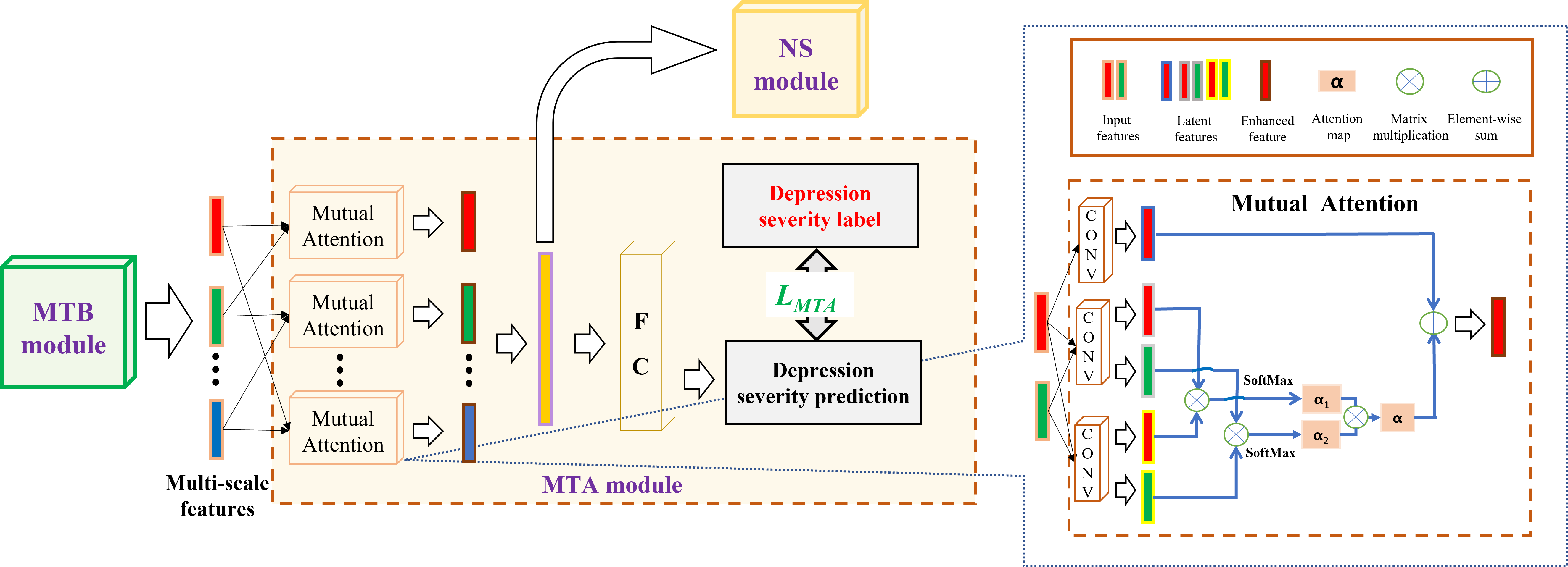}}
	\subfigure[The Noise Separation (NS) module]{\label{subfig:DS}
		\includegraphics[width=16.5cm]{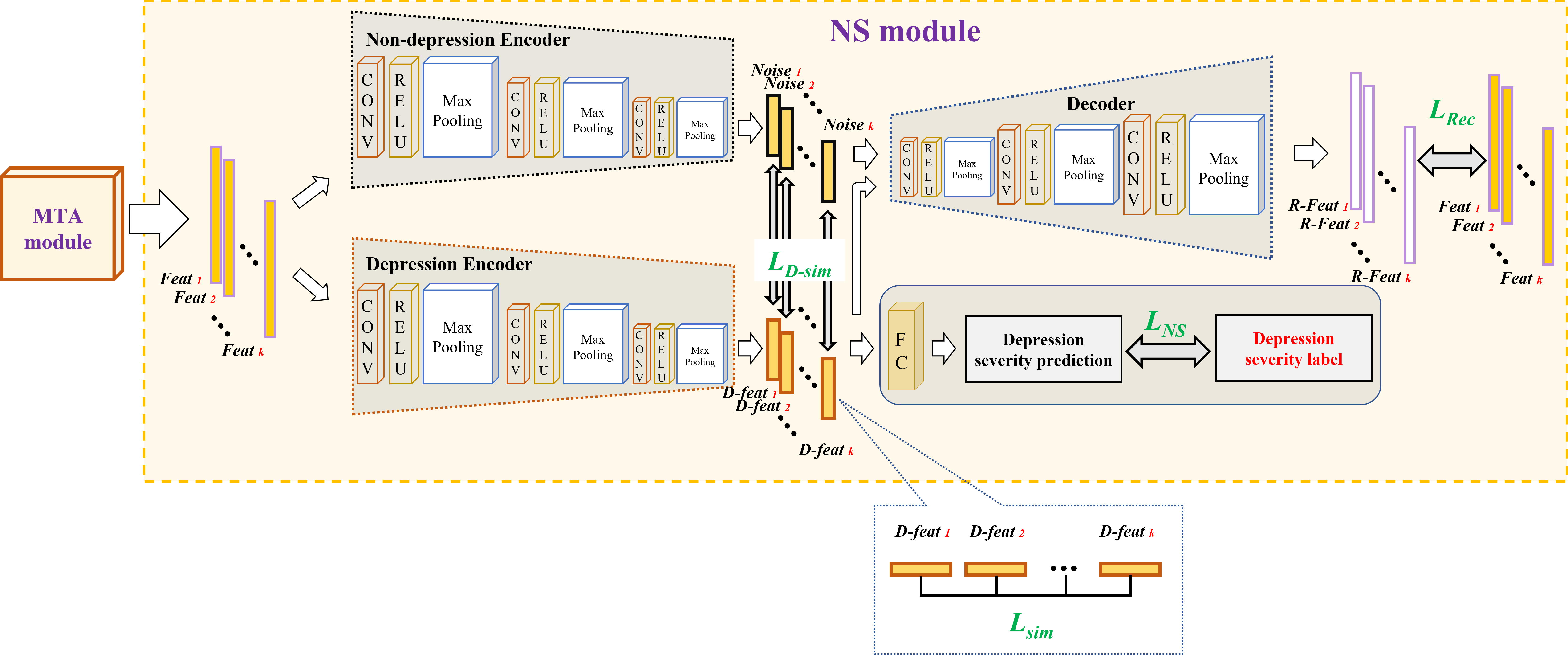}}
	\caption{Illustration of the architectures and training details of the Depression Feature Enhancement (DFE) module.} 
    \label{fig:DFE}
\end{figure*}

\noindent While the proposed MTB module can learn depression-related features at multiple temporal scales, these features may still encode noisy information that is irrelevant to depression recognition. In this paper, we hypothesize that every feature learned from each temporal scale comprises two types of information: depression-related cues and non-depression noise. To further enhance the depression-related cues encoded by the feature whilst removing non-depression noise, we propose a Depression Feature Enhancement (DFE) module. Since the DFE module is designed to be easily plugged on the top of any standard network-based feature extractor, we attach it on the top of our proposed MTB module in this paper. In particular, the DFE module consists of the following two sub-modules:

\textbf{Mutual Temporal Attention (MTA) module:} The main aim of this module is to enhance the depression-related cues encoded by the features learned from each temporal scale, respectively. We hypothesize that the depression-related cues learned at different temporal scales are highly correlated, as all features were learnt to predict the depression severity of the target individual, i.e., predicting the same score. Since the attention operation can explicitly locate and highlight similar semantics between representations, the MTA module aims to identify and enhance the salient regions (the highly correlated information) of all latent features that are learned from MTB. As illustrated in Fig. \ref{subfig:MTA}, the proposed MTA module consists of a set of mutual-attention blocks to identify the underlying relationship between the salient information of each feature pair $f_1^{\text{in}}$ and $f_2^{\text{in}}$, emphasizing the depression-related information of $f_1^{\text{in}}$, i.e., the semantics of $f_1^{\text{in}}$ that highly correlates semantics of $f_2^{\text{in}}$. In particular, both inputs of a mutual attention block are projected to two latent spaces using $1 \times 1$ convolution layers as:
\begin{equation}
   f_1^{L1} = \text{Conv}_\beta(f_1^{\text{in}}),   \quad f_1^{L2} = \text{Conv}_\omega(f_1^{\text{in}})
\end{equation}
\begin{equation}
   f_2^{L1} = \text{Conv}_\beta(f_2^{\text{in}}),  \quad f_2^{L2} = \text{Conv}_\omega(f_2^{\text{in}})
\end{equation}
Then, a matrix multiplication operation is conducted for each feature to compute the similarity between the produced two latent embedding. As a result, two attention maps can be produced as:
\begin{equation}
   f_1^{\text{attention}} = (f_1^{L1})^{\top} (f_1^{L2})
\end{equation}
\begin{equation}
   f_2^{\text{attention}} = (f_2^{L1})^{\top} (f_2^{L2})
\end{equation}
This step is inspired by the non-local attention strategy that captures the global dependencies to highlight the salient information encoded by the corresponding feature. Then, we further conduct the matrix multiplication operation between the attention maps that come from two inputs, in order to further generate an attention map that can enhance the most important depression cues in $f_1^{in}$:
\begin{equation}
   f^{\text{attention}} = (f_1^{\text{attention}})^{\top} (f_2^{\text{attention}})
\end{equation}
As a result, the final enhanced feature $f_1^{\text{MTA}}$ that corresponds to $f_1^{\text{in}}$ can be produced by applying $f^{\text{attention}}$ to weight a latent representation of $f_1^{\text{in}}$: 
\begin{equation}
   f_1^{\text{{MTA}}} = \gamma(f_1^{\text{in}}) \oplus (f^{\text{attention}})
\end{equation}
In summary, the final output $F$ that aggregates all the enhanced features ($f_1^{\text{MTA}}, f_2^{\text{MTA}}, \cdots, f_k^{\text{MTA}}$) should provide more reliable representations for depression severity prediction.

\textbf{Noise Separation (NS) module:} While the proposed MTA module can identify and enhance the depression-related cues, the non-depression noise may still be retained in the generated latent representation. The assumption is that the latent representation generated by MTA is made up of two parts of information: depression-related cues and non-depression noise. Therefore we introduce a Noise Separation (NS) module to disentangle the depression-related information and non-depression noises of the latent feature. In particular, we train a CNN block that takes the feature generated by the MTA as the input and further disentangles it to depression-related and non-depression component. This module is inspired by the approach introduced in \cite{bousmalis2016domain}. During the training stage, as illustrated in Fig. \ref{subfig:DS}, the NS module contains a shared depression feature encoder and a shared non-depression feature encoder, aiming to outputs depression-related features and non-depression features from a set of inputs, respectively. We first assign all videos into four depression categories (these categories are decided based on their BDI II scores), namely, minimal depression, mild depression, moderate depression and severe depression. At each training iteration, we only provide a set of latent features that belong to the same depression category as the inputs to both encoders. We use a regressor attached to the depression encoder, enforcing it to learn features that are relevant to the depression severity estimation. We also enforce feature similarity within the generated depression features by minimizing their difference for the given set of input features with the same depression category. Meanwhile, we maximize the difference between each depression-related feature and its corresponding non-depression feature produced by the non-depression feature encoder. Since the assumption is that each input feature is only made up of corresponding depression-related cues and non-depression noise, a decoder that reconstructs the input features based on both produced depression-related and non-depression features is attached. In this way, the non-depression noises can be specifically attenuated. During the inference stage, we only utilize the features generated by the depression encoder. It would distill only depression-related information and those not pertaining to depression will be removed by the disentanglement process.

\subsubsection{Loss functions for MTB-DFE training}
\label{subsec:loss_func}

\noindent Given that at each training iteration there are $N$ input features corresponding to $N$ video clips of the same depression category, the loss functions for training the MTB-DFE module are explained as follows. 

Firstly, we employ the Mean Square Error (MSE) loss function to measure the difference between the depression severity predictions $p_n^\text{NS}$ generated by the MTB-DFE module (i.e., the output of the NS module) and their corresponding depression severity ground-truth $g_n$ ($n = 1, 2, \cdots, N$), denoted as
\begin{equation}
\label{eq:DS}
L_{\text{NS}} = \frac{1}{N} \sum_{n=1}^{N} \left(p_n^\text{NS} - g_n \right)^{2}
\end{equation}
Then, we attach an auxiliary head to the MTA module for intermediate supervision thereby enforcing the MTA module to predict the depression severity label, where the same MSE loss function is again used (Eq. \ref{eq:MTA}). This method augments the network's capacity to extract depression-related features.
\begin{equation}
L_{\text{MTA}} = \frac{1}{N} \sum_{n=1}^{N} \left(p_n^{\text{MTA}}-g_n\right)^{2}
\label{eq:MTA}
\end{equation}
We adopt three other loss functions besides the aforementioned loss terms during the training of the NS module. Since the objective of the depression encoder is to extract features from video clips of different individuals who have the same depression category at each training iteration, the features extracted from these clips should be very similar. Thus, we define such similarity in terms of:
\begin{equation}
\label{eq:similarity}
L_{\text{sim}} = \frac{1}{N^2}\sum_{n=1}^{N-1} \sum_{i=n+1}^n (F_{n}^D-F_{i}^D)^2
\end{equation}
where $F_{n}^D$ and $F_{i}^D$ are the depression-related features extracted from the shared depression encoder while $n$ and $i$ are the indices of input features that come from the different individuals with the same depression category. This training strategy allows the depression encoder focusing on learning common depression-related short-term facial behaviours from the input clips, which are invariant to the differences in identity, gender, age, etc. 

We then use the $L_{\text{D-sim}}$ loss to encourage depression-related and non-depression feature components extracted from the same clip to be orthogonal (dissimilar), which is defined as
\begin{equation}
L_{\text{D-sim}} = \frac{1}{N^2} \sum_{n=1}^{N} \left\|(F_{n}^{\text{Dep}})^{\top} F_{n}^{\text{Non}} \right\|_{\text{Frob}}^{2}
\end{equation}
where $F_{n}^{\text{Dep}}$ and $F_{n}^{\text{Non}}$ are the depression-related  and non-depression components of the $n_{th}$ input feature. $\|\cdot\|^{2}_F$ is the square Frobenius norm. To further ensure the input feature's disentanglement without losing any crucial information, we introduce a reconstruction loss function (Eq. \ref{eq:rec}) that allows the input of the NS module to be reconstructed from the extracted depression-related and non-depression feature components using the decoder, which we define as:
\begin{equation}
\label{eq:rec}
L_{\text{Rec}} = \frac{1}{N\times J} \sum_{n=1}^{N} \sum_{j=1}^{J} \left(F_n^{\text{Dec}}(j)-F_n(j)\right)^{2}
\end{equation}
where $F_n(j)$ and $F_n^{\text{Dec}}(j)$ are the $j_{th}$ element of the $n_{th}$ input feature and the $j_{th}$ element of the corresponding reconstructed feature generated by the decoder. 

As a result, the final loss function for optimizing the MTB-DFE module can be defined as the combination of the above loss functions:
\begin{equation}
\label{eq:short}
\begin{split}
L_{\text{short}} & =  L_{\text{NS}} + W_1 \times L_{\text{MTA}} + W_2 \times L_{\text{sim}}  \\ 
&  + W_3 \times L_{\text{D-sim}} + W_4 \times L_{\text{Rec}}
\end{split}
\end{equation}
where $W_1$, $W_2$, $W_3$ and $W_4$ represent the importance of each loss, respectively. In this paper, we set all of them as $1$.



\subsection{Video-level depressive behaviour modelling}
\label{subsec:graph_representation}

\begin{figure}
\centering
\includegraphics[width=8.8cm]{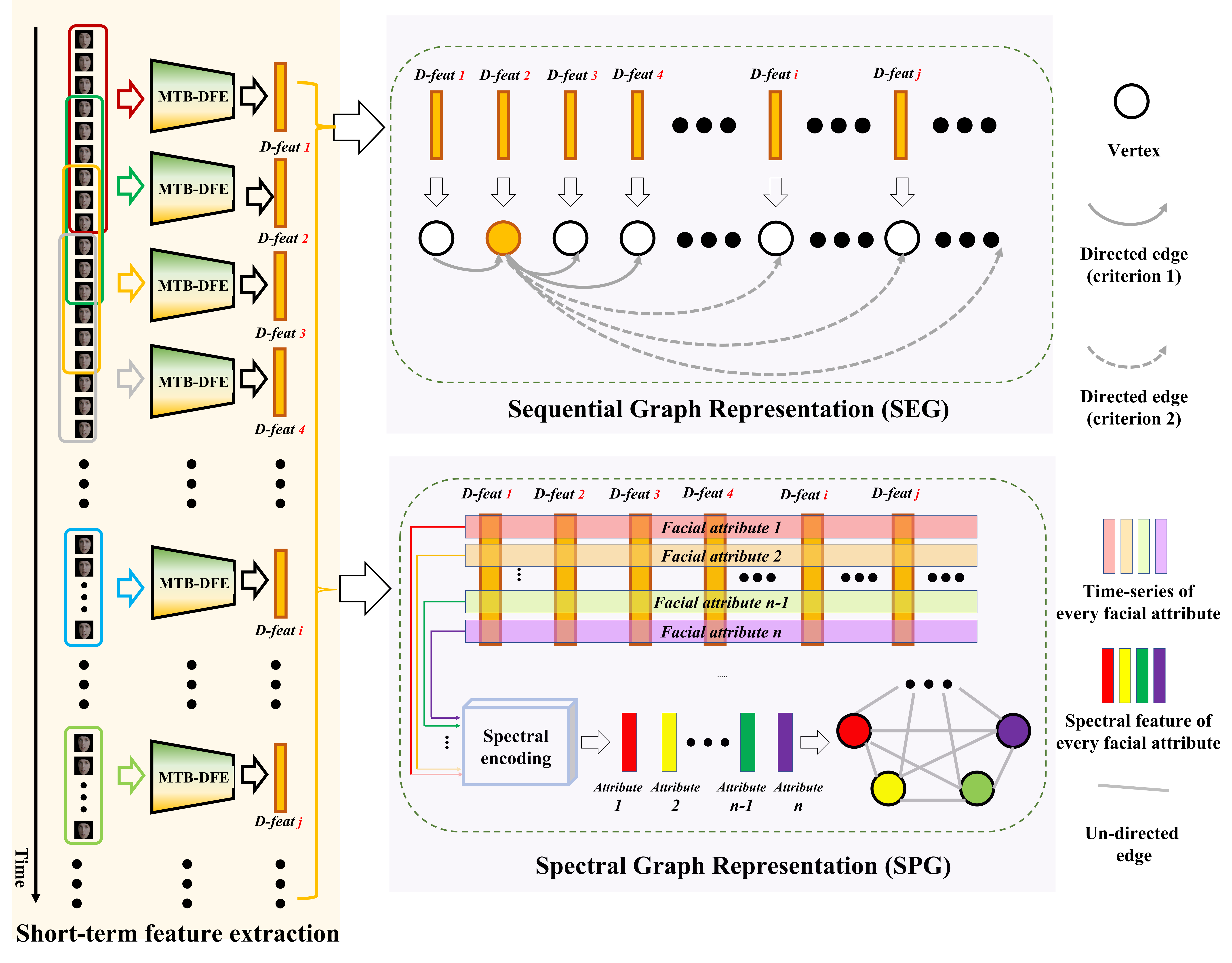}
\caption{Illustration of the proposed two graph representation encoding strategies, where each strategy takes all deep-learned thin slice-level features of the target video (depicted in yellow) as the input and then produce a video-level graph representation (depicted in purple). In SEG encoding module, we only show the edges of the second vertex (the yellow circle).}
\label{fig:graph_encoding}
\end{figure}

\noindent Besides the short-term depression-related facial behavioural cues, long-term behaviours usually act as a more reliable source for estimating depression severity. To this end, we first recall the main issues encountered to construct video-level (long-term) representations for video-based automatic depression analysis: (i) while standard ML/CNN models require the input videos to confirm to a fixed size, face videos collected from different subjects usually have variable lengths; and (ii) the original videos usually contain a large number of frames, which cannot be directly provided to ML/CNN models. Simply computing the statistics (e.g., average values) from all thin video slices' predictions/features \cite{al2018video,zhou2018visually} forgoes key facial dynamics, while down-sampling variable-length videos to the same length \cite{de2021mdn} discards a large number of frames carrying vital information. In order to mitigate these problems, we propose two video-level facial behavioural graph representation encoding strategies: sequential graph representation (SEG) and spectral graph representation (SPG), which not only encode multi-scale short-term and long-term facial dynamics but also retain the information from all available frames of the target video, regardless of its length. Both graph representation encoding strategies are visualized in Fig. \ref{fig:graph_encoding}.

\subsubsection{Sequential graph representation}
\label{subsec:sequential-graph}

\noindent We firstly propose to directly represent the variable-length face videos as Sequential Graph Representations (SEG) which characterize variable numbers of vertices and edges, in order to represent the video-level depression-related facial behaviours of the target subjects. For a SEG, each thin slice-level depression-related feature in a video is represented as a vertex. Each vertex is connected to other vertices based on two criteria, their temporal adjacency in the video and the pre-defined temporal scales. In particular, the first criterion for vertex connectivity requires the vertices representing the thin slices of the same video to have overlapping content (or temporal adjacency). The second criterion allows connections based on a set of pre-defined time-windows. This connectivity setting facilitates SEG to capture facial behaviours at multiple temporal scales. All edges in the SEG are directed, as they display the time flow of the corresponding vertices.

It is important to emphasize that the generated SEGs are length-dependent, (i.e., the number of vertices in SEG equals to the video length). We propose to employ Heterogeneous Graph Neural Network techniques \footnote{\url{https://docs.dgl.ai/en/0.6.x/guide/graph-heterogeneous.html}}to process the resulting variable-size SEGs (heterogeneous graphs). This facilitates the variable-length videos to be directly encoded to predict depression at the video-level using all available frames.

\subsubsection{Spectral graph representation}
\label{subsec:spatial-graph}

\noindent While the SEG is a straightforward approach to construct video-level heterogeneous graph representations, we further propose a spectral graph representation (SPG) that summarizes thin slice-level depression features of an arbitrary length video into a length-independent isomorphic graph representation. In the SPG, we treat each dimension of the thin slice-level features as a vertex, i.e., the number of vertices in an SPG equates to the dimension of the thin slice-level feature. Since we compute the short-term behavioural features from the thin slices of all videos using the same MTE-DFE framework, the dimensions of all thin slice-level features are the same. As a result, the SPGs of all videos would have the same number of vertices, regardless of their lengths.

The SPG is designed to represent the video-level behavioural information, each vertex in a SPG represents the time-series of a facial attribute over all thin slices of the video. However, if we directly use the time-series of each facial attribute as a vertex feature, the dimension of vertices' features for a SPG would match the number of thin slices of the corresponding video, which leads SPGs of variable-length videos to have different vertex feature dimensions. To this end, we extend the spectral encoding algorithm \cite{song2018human,song2020spectral} to individually process the facial attribute time-series, converting facial attribute time-series of each video to a length-independent spectral vector. In particular, the time-series of each facial attribute of each video is first transformed to a spectral signal using Discrete Fourier Transform, where the number of frequencies (the dimension) of the spectral signal equates to the number of thin slices of the corresponding video. Since the difference in videos' lengths would lead the produced spectral signals to have different frequency components, we choose the common frequencies comprised by spectral signals of all videos. Subsequently, all spectral signals would have the same dimension corresponding to the same set of frequencies. Finally, we select the Top-K low-frequency components as the vertex feature for each facial attribute, as the low-frequency components usually encode the most important cues (please see \cite{song2020spectral} for details). As a result, all SPGs would have $K$-dimension vertices' features regardless of their videos' lengths. In other words, assuming that the MTB-DFE extracts $M$ facial attributes ($M$-D short-term feature) from each thin slice, for a video with an arbitrary number of frames, we construct an SPG that has $M$ vertices, where each of them has $K$ dimensions. More importantly, each dimension in the vertices' features corresponds to a unique video-level frequency representing how fast the corresponding facial attribute changes in time. Consequently, each vertex feature in the SPG contains multi-scale ($K$ temporal scales) video-level dynamics of the corresponding depression-related facial behaviour attributes. 

We present a more flexible and elaborate approach to video-level representation learning in comparison to the original spectral vector introduced in \cite{song2018human,song2020spectral}, where they simply concatenate the spectral features of all attributes as a one-dimensional vector. This rigid approach disregards the properties of the spectral components of the features and treats all spectral dimensions of all channels equally. The concatenation operation does not take into account whether two features correspond to the same frequency or share the same channel, losing important discriminative information encoded by the spectral representation. However in the proposed novel SPG representation, all spectral features corresponding to a given channel are assigned to an independent vertex and each dimension of the vertex represents a given frequency. Therefore, the SPG provides a higher representational capability compared to the original spectral vector.

\subsection{Depression recognition}
\label{subsec:GCN}

\noindent Once the video-level graph representation is obtained, we employ the state-of-the-art Graph Attention Network (GAT) \cite{velivckovic2017graph} to predict depression severity. The GAT uses masked self-attention layers to assign different weights for various vertices. Importantly, it can simultaneously process graphs with different architectures. In this paper, the GAT model is made up of GAT layers and fully connected (FC) layers in order to output a single depression severity score from each input graph representation.

\section{Experiments}

\noindent In this section, we first provide the details of the AVEC 2013 and AVEC 2014 audio-visual depression datasets that are used for evaluating the proposed approaches (Sec. \ref{subsec:dataset}). Then, the implementation details, including data pre-processing, the settings of short-term and video-level feature extraction models, the depression recognition model (GAT), training details, and evaluation metrics are detailed in Sec. \ref{subsec:Implementation details}. Subsequently, Sec. \ref{subsec:sota} compares the proposed approach with other recently proposed methods. In addition, we present a set of ablation studies in Sec. \ref{subsec:ablation} that aims to investigate the influence of various settings on depression severity prediction performance, including: (i). multi-scale short-term facial behaviour temporal modelling; (ii). the proposed Depression Feature Enhancement module; (iii). the video-level graph representations; and (iv). the proposed two-stage framework. Finally, we report the cross-dataset evaluation in Sec. \ref{subsec:cross-dataset}.

\subsection{Datasets}
\label{subsec:dataset}

\noindent Our experiments were conducted on the audio-visual depression corpus corresponding to AVEC 2013 \cite{valstar2013avec} and AVEC 2014 \cite{valstar2014avec} challenges. The corpus used by the AVEC 2013 challenge contains $150$ audio-visual clips, where each clip records a subject engaging in a set of pre-defined tasks, e.g., speaking out loud while solving a task, sustained vowel phonation, sustained loud vowel phonation, counting from 1 to 10, and sustained smiling vowel phonation. The duration of AVEC 2013 videos ranges from $20$ minutes to $50$ minutes with an average of $25$ minutes. The corpus used by the AVEC 2014 challenge also contains $150$ audio-visual clips, where each clip contains two sub-clips that individually record two tasks: Northwind and Freeform. In comparison to AVEC 2013 corpus, the duration of the sub-clips in AVEC 2014 are much shorter (ranging from $6$ seconds to $4$ minutes $8$ seconds). For both datasets, each clip is labeled with a Beck-Depression Inventory (BDI II) score indicating a depression severity that ranges from a minimum of $0$ to a maximum of $63$.


\subsection{Implementation details}
\label{subsec:Implementation details}

\subsubsection{Video pre-processing}

\noindent In our experiments, the face region of each frame is cropped and aligned using OpenFace 2.0 \cite{baltrusaitis2018openface} based on the CE-CLM landmark detector, where the resolution of the obtained face image is $112 \times 112$. For each frame where the face detection fails, we replace it with the face image extracted from the nearest frame in the video before the model training.

\subsubsection{Model settings}


\noindent \textbf{MTB module:} In this paper, we employ the MTB module consisting of three ResNet-50 networks which were pre-trained on VGGFace2 \cite{cao2018vggface2}. It provides $256$, $512$ and $2048$ feature maps with the sizes, $8 \times 28 \times 28$, $8 \times 14 \times 14$, $8 \times 4 \times 4$, respectively. The final output of the MTB module comprises three temporal feature map sets, each of which consists of $1024$ feature maps with the size of $1 \times 4 \times 4$. Finally, each feature map set is converted to a 1D latent feature vector of $2048$ dimensionality thereby forming the input for the DFE module.

\textbf{DFE module:} The DFE module is made up of an MTA module and an NS module. As illustrated in Fig. \ref{subfig:MTA}, the MTA module consists of three non-local modules to independently capture the salient information of each temporal scale as well as three mutual attention modules that enhance the correlated information from each of the feature pairs. The NS module is a standard encoder that contains four 1-D convolution layers with $1024$, $512$, $128$ and $32$ kernels. During the training of the NS module, the shared non-depression encoder has the same architecture as the depression encoder, i.e., it also generates a $32$-D non-depression feature vector for each input. The decoder used for feature reconstruction consists of three 1D convolution layers with $128$, $512$ and $2048$ kernels, respectively, while the depression regressor is an FC layer with ReLU activation function. In the NS module, the kernel size of all convolution layers is set to $1$.

\textbf{Depression recognition model:} In this paper, the employed GAT model contains one GAT layer, a readout layer and three FC layers with ReLU activation function attached. In particular, we adopted the “mean” operation to aggregate the nodes' features in the readout layer.

\subsubsection{Training details}

\noindent We conducted standard training, validation and testing using the training, validation and test data provided by each dataset (AVEC 2013, AVEC 2014 NorthWind, and AVEC 2014 Freeform), respectively. During the training of the MTB-DFE module, we set the batch size to $5$ thin slices, where each slice consists of $30$ consecutive frames. The Adam \cite{kingma2014adam} optimizer is employed to optimize the MTB-DFE framework. The training of MTB-DFE module is achieved by jointly minimizing the a set of corresponding loss functions (explained in Sec. \ref{subsec:loss_func}), where the $W_1$, $W_2$ and $W_3$ in Eq. \ref{eq:short} are all set to $1$ in this paper. To train the GAT, we set the batch size to $1$. The Adam optimizer is utilized with MSE as the loss function. It should be noted that for each dataset, we kept the hyper-parameters consistent for all experiments. All hyper-parameters used in this paper are detailed in the supplementary document. Besides the spectral representation which was implemented in MATLAB, all other experiments were implemented in the PyTorch library while the DGL library was used for building GNNs. 

\subsubsection{Evaluation metrics}

\noindent Four metrics used by previous AVEC challenges \cite{valstar2013avec,valstar2014avec,ringeval2019avec} are employed to compare the performance of the proposed approach. Firstly, Root Mean Square Error (RMSE) and Mean Absolute Error (MAE) are introduced to measure the errors between the predictions and ground-truth, which are defined as:
\begin{equation}
\mathrm{RMSE}=\sqrt{\frac{1}{n} \sum_{i=1}^{n}\left(p_{i}-g_{i}\right)^{2}}
\end{equation}
\begin{equation}
\mathrm{MAE} = \frac{1}{n} \sum_{i=1}^{n} \left|p_i-g_i
\right|
\end{equation}
where $p_i$ is the $i^{th}$ depression severity prediction and $g_i$ is the $i^{th}$ ground-truth. In addition, we also report two metrics for correlation between the predictions and the ground-truth based on the Pearson Correlation Coefficient (PCC) and the Concordance Correlation Coefficient (CCC). PCC measures the linear correlation between the predictions $P$ and their corresponding ground-truth $G$:
\begin{equation}
\mathrm{PCC}=\frac{cov(P, G)}{\sigma_{P} \sigma_{G}}
\end{equation}
where $cov(P, G)$ is the co-variance function; $\sigma_{P}$ and $\sigma_{G}$ are the standard deviations of $P$ and $G$. The CCC is employed to measure the reproducibility/inter-rater reliability between the predictions $P$ and their corresponding ground-truth $G$, which is defined as
\begin{equation}
\rho_{c}=\frac{2 \rho_{P,G} \sigma_{P} \sigma_{G}}{\sigma_{P}^{2}+\sigma_{G}^{2}+\left(\mu_{P}-\mu_{G}\right)^{2}}
\end{equation}
where $\rho_{P,G}$ is the PCC between $P$ and $G$; $\mu_{P}$ and $\mu_{G}$ are the mean values of the predictions and ground-truth, respectively; $\sigma_{P}$ and $\sigma_{G}$ are the corresponding standard deviations.

\subsection{Comparison to existing approaches}
\label{subsec:sota}

\noindent In this section, we compare the proposed approach to the existing state-of-the-art methods. According to Table \ref{tb:soat_avec2013} and Table \ref{tb:soat_avec2014}, the proposed short-term modelling module (MTB-DFE model) already attains the second best performance among all listed thin slice-based depression recognition methods, which is comparable to the state-of-the-art \cite{de2020deep}. These results demonstrate the competitiveness of the proposed MTB-DFE module and its superiority among approaches that model depression from thin video slices, showcasing its strong capacity to capture depression-related short-term facial behavioural cues. Importantly, the proposed DFE module is versatile and can be easily plugged into most existing deep learning frameworks (analysed in Sec. \ref{subsec:ab-short-term}).

Both of the proposed video-level depression graph representations achieve promising results that demonstrate large performance gains over most of the existing video-level depression modelling approaches. The SPG-based two-stage framework surpasses all of the listed video-level modelling approaches with $5\%$ RMSE improvements over the previous state-of-the-art method \cite{de2021mdn} on AVEC 2014 datasets. We hypothesize that while \cite{de2021mdn} and \cite{de2020deep} can provide reliable predictions for subjects' depression status based on either long-term or short-term facial behaviours, the proposed two-stage framework can specifically model depression by incorporating both long-term and short-term behaviours. As a result, it achieves superior performance over most of the existing one-stage approaches. Notably, we found that the decision-level fusion of the predictions obtained from both tasks of the AVEC 2014 dataset can provide better predictions for all methods (\cite{de2020deep,zhou2020depression,de2021mdn} and ours), showing that behaviours triggered by different tasks may contain different but informative cues for depression recognition.

\begin{table}[t]
	\begin{center}
		\begin{tabular}{|l|  c | c | c |}
			\toprule
&Method & MAE & RMSE\\   
\hline \hline
\multirow{8}*{FTL} 

&Baseline \cite{valstar2013avec} & 10.88 & 13.61  \\
&Zhu et al. \cite{zhu2017automated} & 7.58 & 9.82 \\
&Jazaery et al. \cite{al2018video} & 7.37 & 9.28 \\
&Zhou et al. \cite{zhou2018visually} & [6.20] &8.28 \\
&Zhou et al.\cite{zhou2020depression}  & 6.63 &  8.37\\
&Uddin et al. \cite{uddin2020depression} & 7.04 &  8.93\\
&He et al. \cite{he2021automatic} & 6.51 &  8.30\\
&Melo et al. \cite{de2020deep} & \textbf{5.98} &  \textbf{7.90}\\
&Ours (MTB-DFE)  & 6.31 & [8.20]  \\

\hline
\multirow{5}*{VL} 

&Meng et al.\cite{meng2013depression} & 9.14 & 11.19  \\
&Wen et al. \cite{wen2015automated} & 8.22 & 10.27  \\
&Niu et al. \cite{niu2020multimodal}  & 7.32 &  8.97\\
&Song et al. \cite{song2020spectral} & 6.40 &  8.26\\
&Melo et al. \cite{de2021mdn} & [6.06] &  \textbf{7.55}\\
&Ours (MTB-DFE+SEG)  & 6.05 & 7.92   \\
&Ours (MTB-DFE+SPG)  & \textbf{5.95} & [7.57] \\

			\bottomrule
		\end{tabular}
        
	\end{center}
	\caption{Comparison between our systems and other works on \textbf{AVEC 2013} test set, where \textbf{FTL} and \textbf{VL} represent the frame/thin slice-level depression modelling approaches and video-level depression modelling; \textbf{MTB-DFE} denotes the proposed short-term modelling module; \textbf{SEG} and \textbf{SPG} denote the video-level sequential graph representation and spectral graph representation, respectively.} 
\label{tb:soat_avec2013}
\end{table}

\begin{table}[t]
	\begin{center}
		\begin{tabular}{|l|  c | c | c |}
			\toprule
&Method & MAE & RMSE\\   
\hline \hline
\multirow{9}*{FTL} 

&Baseline \cite{valstar2014avec} & 8.86 & 10.86  \\
&Sidorov et al. \cite{sidorov2014emotion} & 11.20 & 13.87 \\
&Zhu et al. \cite{zhu2017automated} & 7.47 & 9.55 \\
&Jazaery et al. \cite{al2018video} & 7.22 & 9.20 \\
&Jan et al. \cite{jan2017artificial}  &6.68 & [8.01] \\
&Zhou et al. \cite{zhou2018visually} &\textbf{6.21} &8.39 \\
&Zhou et al. \cite{zhou2020depression}  & 6.59 &  8.30\\
&Uddin et al. \cite{uddin2020depression} & 6.86 &  8.78\\
&He et al. \cite{he2021automatic} & 6.59 &  8.39\\
&Ours (MTB-DFE)  & [6.30]  & \textbf{7.83}  \\

\hline
\multirow{4}*{VL} 

&Niu et al. \cite{niu2020multimodal}  & 6.43 &  8.60\\
&Song et al. \cite{song2020spectral} & 6.78 &  8.30\\
&Melo et al.  \cite{de2020deep} & 6.59 &  8.31 \\
&Melo et al. \cite{de2021mdn} & \textbf{6.06} &  \textbf{7.65}\\
&Ours (MTB-DFE+SEG)  & 6.35 & 7.72  \\
&Ours (MTB-DFE+SPG)  & [6.24] & \textbf{7.65} \\

\hline
\multirow{4}*{CB} 

&Melo et al. \cite{de2021mdn} & [6.24] &  [7.55]\\
&Ours (MTB-DFE)  & 6.36   &  8.04 \\
&Ours (MTB-DFE+SEG) & [6.24] & 7.72  \\ 
&Ours (MTB-DFE+SPG) & \textbf{5.86} &  \textbf{7.18} \\

			\bottomrule
		\end{tabular}
        
	\end{center}
	\caption{Comparison between our systems and other works on \textbf{AVEC 2014} test set, where \textbf{CB} represents depression modelling approaches that provide the final prediction by combining the predictions achieved on NorthWind and Freeform tasks.} 
\label{tb:soat_avec2014}
\end{table}

\begin{figure}
	\centering
	\subfigure[AVEC 2013]{\label{sca_2013}
		\includegraphics[width=6.8cm]{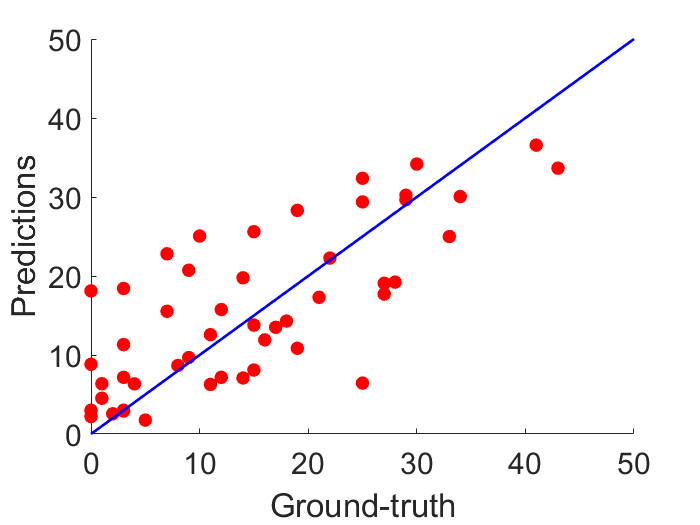}}
	\subfigure[AVEC 2014]{\label{sca_2014}
		\includegraphics[width=6.8cm]{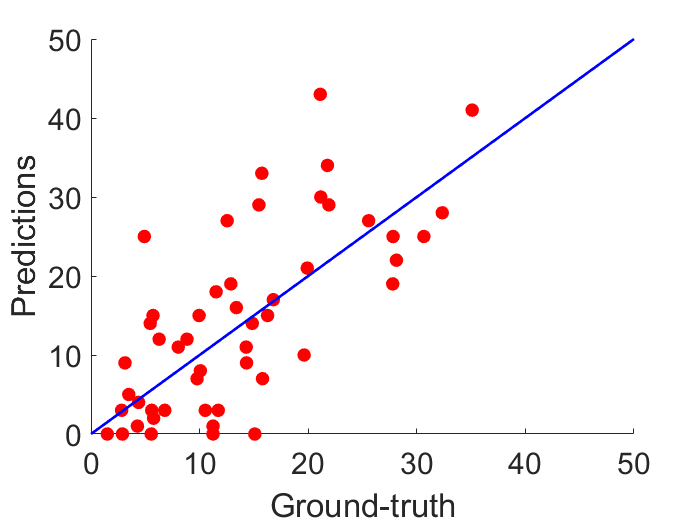}}
	\caption{Predictions of our best system (MTB-DFE+SPG) on AVEC 2013 (top) and AVEC 2014 (bottom) datasets} 
    \label{fig: scatter}
\end{figure}

\subsection{Ablation studies}
\label{subsec:ablation}


\noindent This section explicitly investigates the influence of each of the modules on the proposed two-stage approach, providing evidence and a detailed explanation for the generated state-of-the-art results. All experiments were conducted on the AVEC 2014 Freeform dataset, as this dataset displays spontaneous behaviours of participants, which is closer to real-world scenarios.

\subsubsection{Short-term depression modelling}
\label{subsec:ab-short-term}





\begin{figure}
\centering
\includegraphics[width=8.8cm]{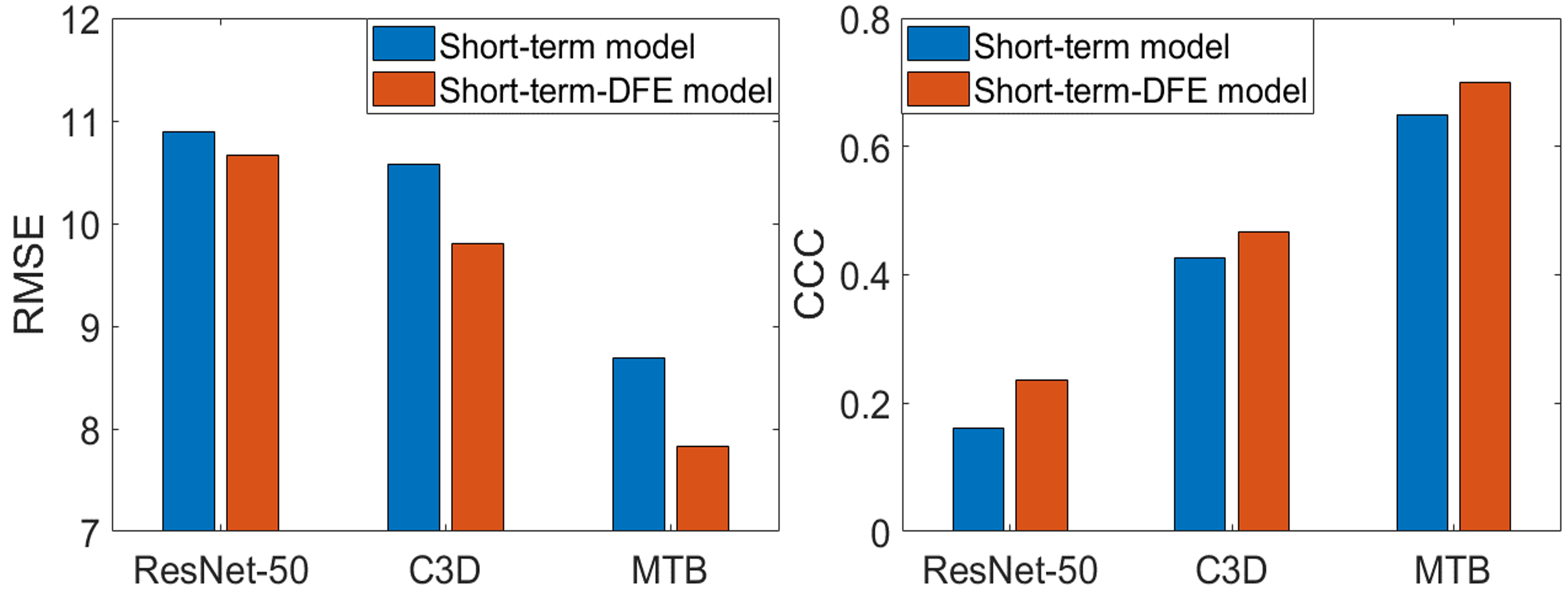}
\caption{Comparison of the results achieved by short-term depression models and their DFE models on AVEC 2014 Freeform dataset.}
\label{fig:result_short-term}
\end{figure}

\noindent We first investigate the advantage of the proposed MTB-DFE module in modelling depression-related short-term facial behaviours. Let's recall from section \ref{subsec:TPAN} that the MTB-DFE module consists of a MTB network that extracts a multi-scale behavioural feature from each thin video slice, as well as a DFE module that consists of a MTA block to enhance the depression-related cues, and a NS block to disentangle the non-depression noise.

Fig. \ref{fig:result_short-term} firstly compares the proposed MTB module to a standard frame-level model (ResNet-50 \cite{he2016deep}) and a single-scale short-term temporal model (C3D network \cite{tran2015learning}), for short-term facial behaviour-based depression recognition. With the same pre-processing settings, the only difference between these three methods is the temporal scale of the extracted features, i.e. ResNet-50 (static feature), C3D (single-scale dynamic feature), and MTB (multi-scale dynamic feature). It can be observed that the proposed MTB achieved better results than both single-scale temporal model and frame-level model, with $17.9\%$ and $20.2\%$ RMSE improvements and  $303.7\%$ and $52.6\%$ CCC improvements, respectively, showing that the depression-related cues are embedded in facial behaviours of multiple temporal scales, (i.e., multi-scale temporal modelling is crucial for face-based depression recognition).

Individually adding the MTA can provide a clear improvement over the MTB module, i.e., MTB-MTA (RMSE $= 8.11$, CCC $=0.693$) achieved $6.7\%$ CCC improvement and $6.2\%$ RMSE improvement over the MTB, which validates the usefulness of the MTA module. Moreover, adding NS module can further enhance the depression recognition performance, with the entire DFE module bringing $7.7\%$ CCC improvement and $10\%$ RMSE improvement to the MTB module. We hypothesize from these results that the proposed DFE module can disentangle the feature representations thereby enhancing the depression-related features and removing the non-depression related noise. In particular, the MTA and NS modules influence different aspects of the input feature, i.e., depression-related cues and non-depression noises, respectively, therefore combining them by a simple concatenation can largely enhance the informative capability of the produced feature.

\begin{figure}
\centering
\includegraphics[width=8.6cm]{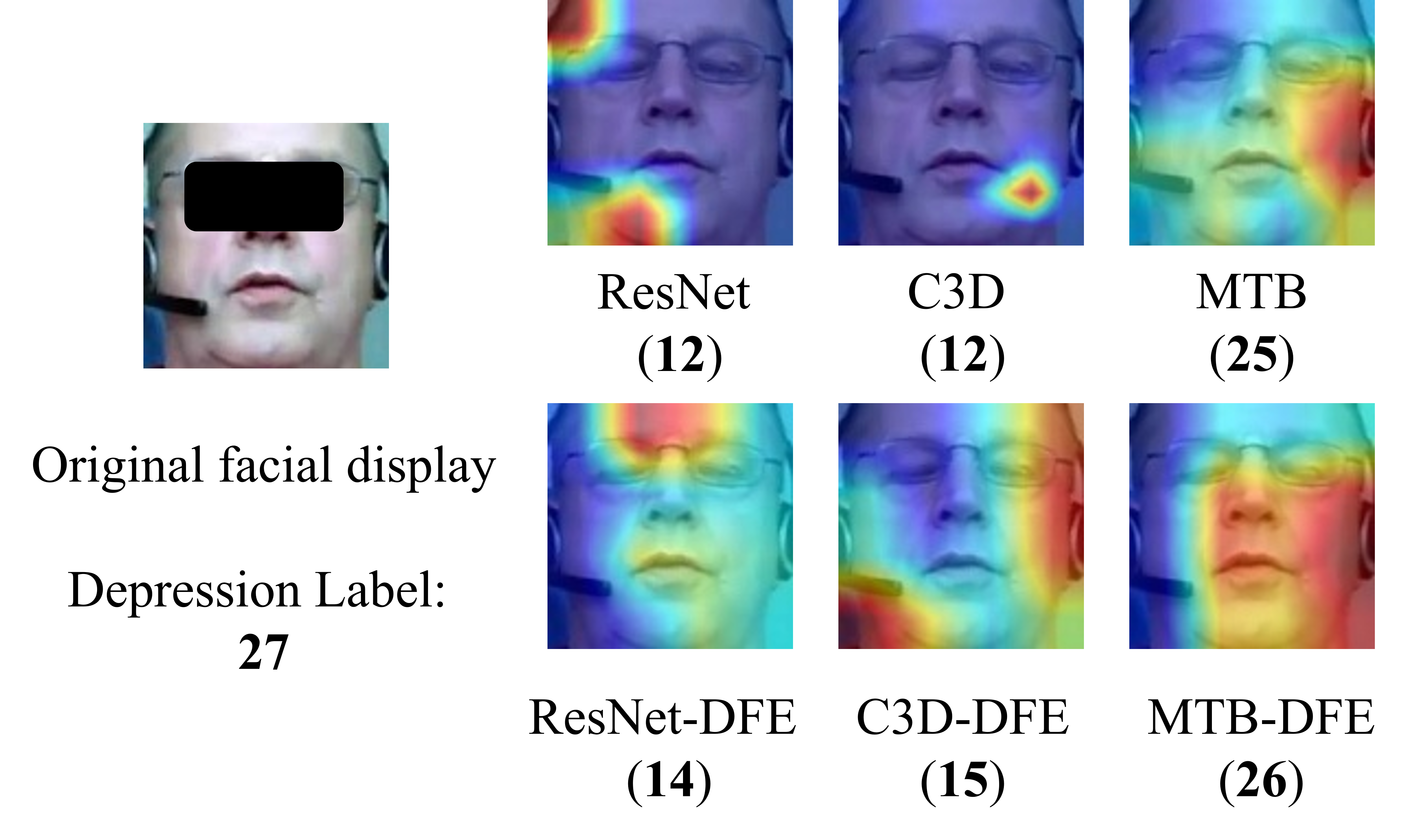}
\caption{Visualisation of the depression-related local facial behaviours for short-term systems and their DFE systems. The number in each bracket denotes the depression prediction achieved from the thin slice of the displayed image.}
\label{fig:visulization}
\end{figure}

To further validate this hypothesis, we also attach the DFE module to ResNet-50 and C3D-based frameworks. Fig. \ref{fig:result_short-term} also clearly shows that the use of DFE can further enhance the short-term facial behaviour-based depression modelling performance. It can be noted that the improvement on ResNet-50 is not as large as the improvements on MTB and C3D models. This may be caused by the fact that the ResNet-50 model only learns depression cues from a static face, and the learned cues may not be reliable (evidenced by poor performance in RMSE and CCC). Therefore, the disentangled ResNet-DFE features still provide limited clues for depression recognition. In addition, we visualize the impact of using the DFE module in Fig. \ref{fig:visulization}. It is clear that the DFE module allows the CNN model to take into account depression-related cues from the facial behaviours of larger facial regions.

\subsubsection{Long-term depression modelling}
\label{subsec:ab-long-term}

\noindent In this section, we investigate the advantages of the proposed graph-based video-level modelling approach. Based on the predictions and latent features generated by the MTB-DFE module, we implemented the following video-level depression severity prediction strategies:

\begin{itemize}
    
    \item \textbf{ATP}: We average all thin slice-level predictions as the video-level prediction \cite{zhou2018visually,al2018video,zhu2017automated}.
    
    
    \item \textbf{STA}: We use $12$ statistics introduced in \cite{song2018human} to represent the video-level information of each feature dimension, and then concatenate the statistics of all feature dimensions as the video-level representation. The produced video-level representation is then fed to a MLP for generating the video-level prediction.
    
    
    \item \textbf{SPV}: We employ the spectral encoding algorithm introduced in \cite{song2020spectral} to summarize all thin slice-level features as a video-level \textbf{spectral vector} which is then fed to a MLP for generating the video-level prediction. 

    \item \textbf{SPH}: We employ the spectral encoding algorithm introduced in \cite{song2020spectral} to summarize all thin slice-level features as a video-level \textbf{spectral heatmap} which is then fed to a 1D-CNN for generating the video-level prediction.
    
    
    \item \textbf{SEG}: We use the proposed sequential graph representation to summarize all thin slice-level features as a video-level representation which is then fed to a 1D CNN for generating the video-level prediction. 
    
    \item \textbf{SPG}: We use the proposed spectral graph representation to summarize all thin slice-level features as a video-level representation which is then fed to the GatedGCN for generating the video-level prediction.
    
\end{itemize}


\begin{figure}
\centering
\includegraphics[width=8.6cm]{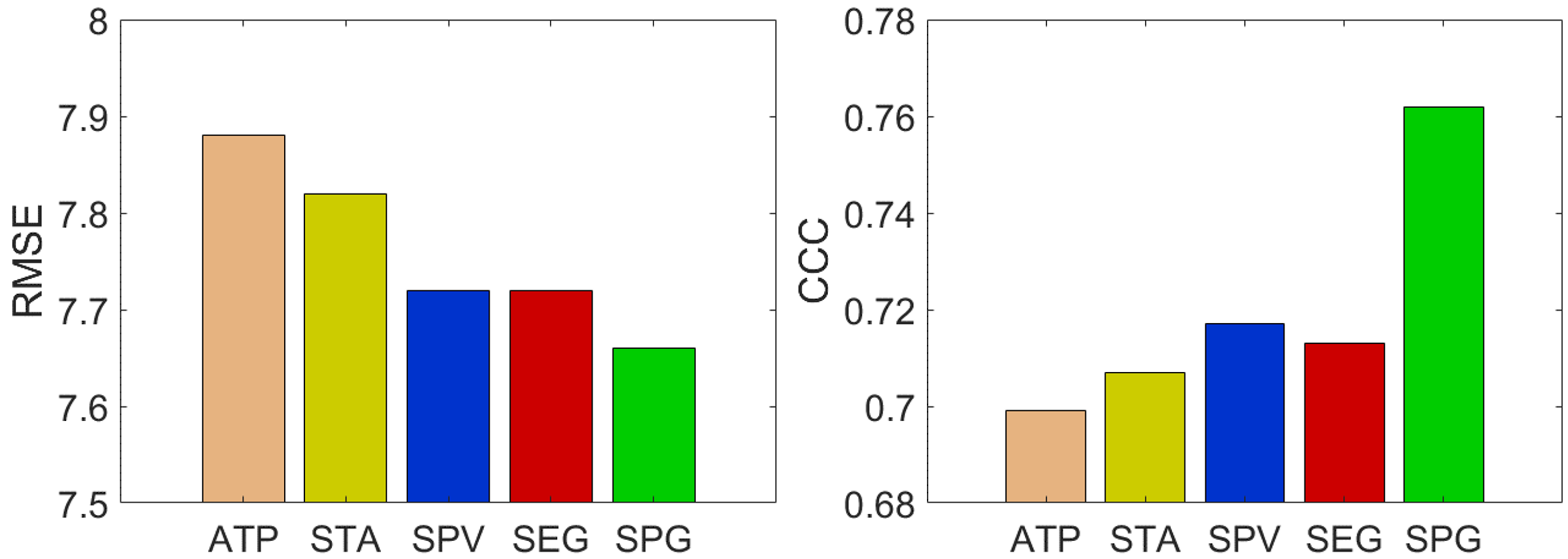}
\caption{Comparison of video-level depression modelling results obtained on the AVEC 2014 Freeform dataset (where the SPH achieved CCC result of $0.383$ and RMSE result of $9.54$).}
\label{fig:result_long-term}
\end{figure}

As illustrated in Fig. \ref{fig:result_long-term}, in comparison to other settings, simply averaging thin slice-level depression predictions or latent features did not provide good results. This may be explained by the fact that despite such strategies computing video-level predictions, those video-level predictions/representations fail to consider temporal dependencies between frames/thin slices, which are crucial for representing depression-related facial behaviours. Among these modelling methods, the STA achieved the better performance as it averages the latent features, which contains more cues than the average of frame-level predictions. In terms of models that encode temporal information, it is clear that the proposed SPG outperforms the SPV, SPH \cite{song2020spectral} and SEG. In particular, the SPV, SPH and SPG use the same spectral feature sets but represent them in distinct ways. The SPV simply concatenates the spectral features obtained from all channels of the produced time-series from the thin slices-level features over the entire video. This approach, as evident, forgoes the channel information. On the other hand, both SPH and SPG can fully address the aforementioned issues, as they represent the spectral features of each channel in an independent channel/vertex and concatenates the spectral features of all channels in a heatmap/graph. Thus, both SPH and SPG preserve the important channel information in a spectral representation. However, we observed that the SPH experiments even generated worse results than that of SPV, which may have been caused by the limited amount of training data. Since existing public audio-visual depression datasets generally contain small number of training data (less than $200$) while GNNs usually are light-weight, characterized by their significantly lower number of parameters for optimization, representing spectral features of all channels as a graph (SPG) is evidently a better way.

Even though using the proposed SEG to model depression at the video-level improved the results compared to the proposed MTB-DFE module, the SEG setting is not as good compared to the SPG. One reason could be that the lengths of AVEC 2014 Freeform videos vary a lot, i.e., the longest video has 7440 (270 thin slices) frames while the shortest video only contains 180 frames (31 thin slices). Consequently, there are large differences in the sizes of the produced SEGs. While the video length should not contribute to the depression severity assessment, this factor can heavily influence the GAT processing procedure as it determines the size and topology of the produced SEG.

\subsubsection{Analysis of the two-stage depression modelling strategy}

\noindent Since the proposed approach establishes the promising results on both datasets and the performance of the proposed multi-scale short-term (MTB-DFE) and video-level (SPG) modelling approaches were validated in previous sections, we now specifically investigate the advantages of the proposed two-stage framework in this section.

We implement a set of short-term and video-level modelling approaches, and then integrate them into the proposed two-stage frameworks. More specifically, we implement four short-term models to extract four types of short-term facial behaviour descriptor for each frame/thin slice, which are: OpenFace 2.0 \cite{baltrusaitis2018openface} that provides frame-level AUs, gaze and head pose ($29$ attributes are used in \cite{song2020spectral}),  ResNet-50 \cite{he2016deep} that learns deep, frame-level depression-related facial features, C3D network \cite{tran2015learning} that deep learns short-term depression-related facial features, and the proposed MTB-DFE that deep learns multi-scale enhanced short-term depression-related facial features. C3D and MTB-DFE are also individually employed as video-level models by down-sampling each target video into a certain number of frames ($30$ frames in this paper) which are then fed to the model for video-level feature learning and depression recognition. Finally, we implement the two-stage framework by using four types of video-level encoding strategies for summarising short-term features, including the ATP, STA, SPV and SPG described in Sec. \ref{subsec:ab-long-term}.

\begin{figure}
	\centering
	\subfigure[The average results achieved for each short-term models.]{\label{subfig:two_short}
		\includegraphics[width=8.6cm]{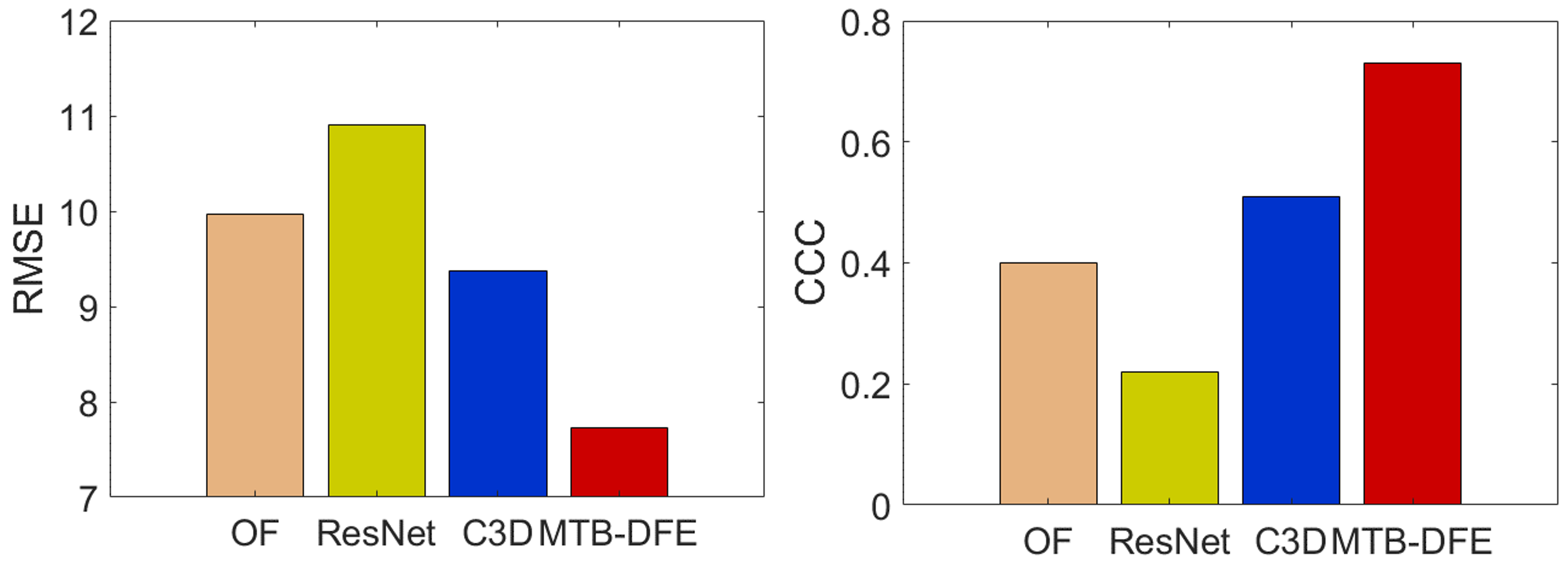}}
	\subfigure[The average results achieved for each long-term models.]{\label{subfig:two_long}
		\includegraphics[width=8.6cm]{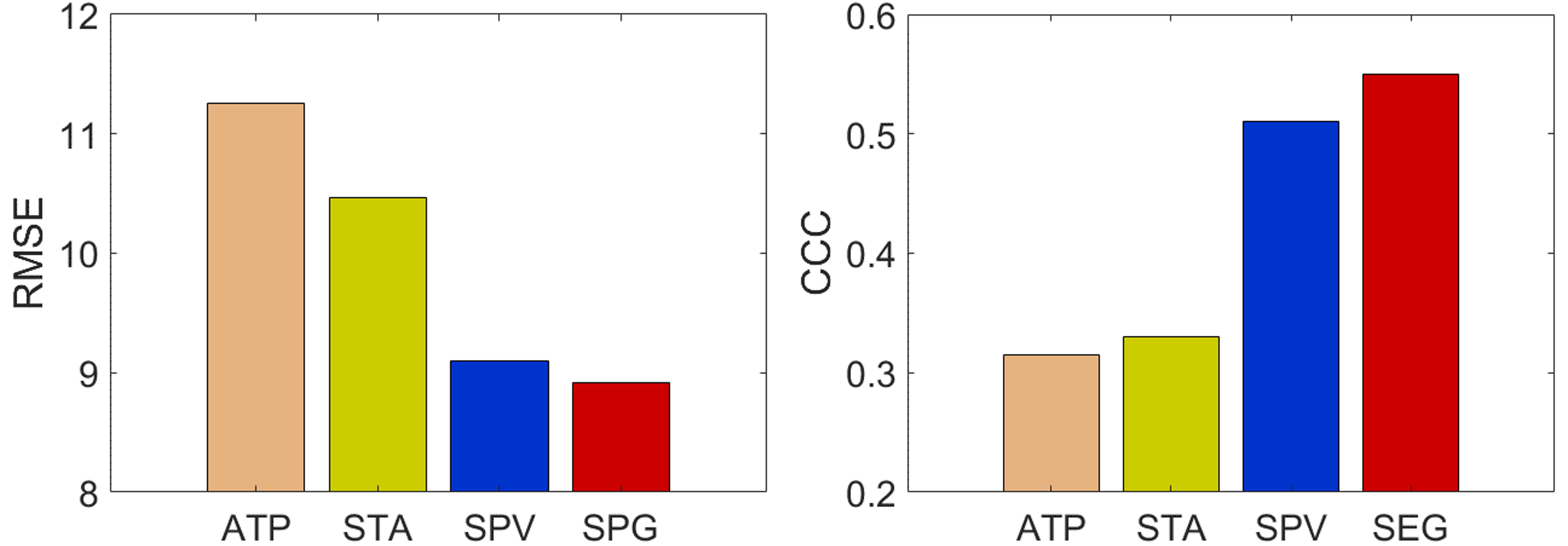}}
	\caption{Comparison of short-term and video-level models on AVEC 2014 Freeform dataset, where OF denotes that the frame-level features are extracted using OpenFace 2.0. Each displayed number is the average result achieved by combining a short-term/long-term model with all long-term/short-term models.} 
    \label{fig:result_two_stage}
\end{figure}

Fig. \ref{subfig:two_short} compares the average results achieved by four short-term models when combining them with video-level encoding strategies, i.e., first extracting all frame/thin slice-level descriptors of the target video and then fusing them as a video-level representation for depression recognition. It is clear that the ATP setting that simply averages the frame/thin slice-level predictions without any specific video-level encoding shows the worst performance. On the contrary, the SPG and SPV yield the most promising results, providing an average of $61.9\%$ and $74.6\%$ average CCC improvements as well as $19.2\%$ and $20.8\%$ average RMSE improvements over all short-term models (ATP). This is because both of them consider multi-scale video-level facial dynamics. These results validate that a proper video-level encoding can provide large and additional performance improvements to short-term models for video-based depression recognition. This can be explained by the fact that long-term behaviour cues are crucial for video-based depression analysis, as people with different depression status can display similar short-term behaviours \cite{song2020spectral}.

Fig. \ref{subfig:two_long} compares the average results achieved by four video-level models when combining them with short-term models. It can be observed that the differences in short-term models also caused large differences in the final depression recognition results, i.e., the short-term models with better performance allow the corresponding two-stage frameworks to also achieve better recognition results, where the MTB-DFE achieved the best results and the OpenFace achieved the worst performance as MTB-DFE deep learns multi-scale enhanced short-term depression-related facial features while OpenFace only extracts mid-level facial attributes without specifically considering the depression-related cues. In other words, a proper short-term model can extract more reliable and depression-related short-term behaviour cues from the original video data, which further allows the two-stage framework to construct a better video-level depression representation.

\begin{figure}
\centering
\includegraphics[width=8.8cm]{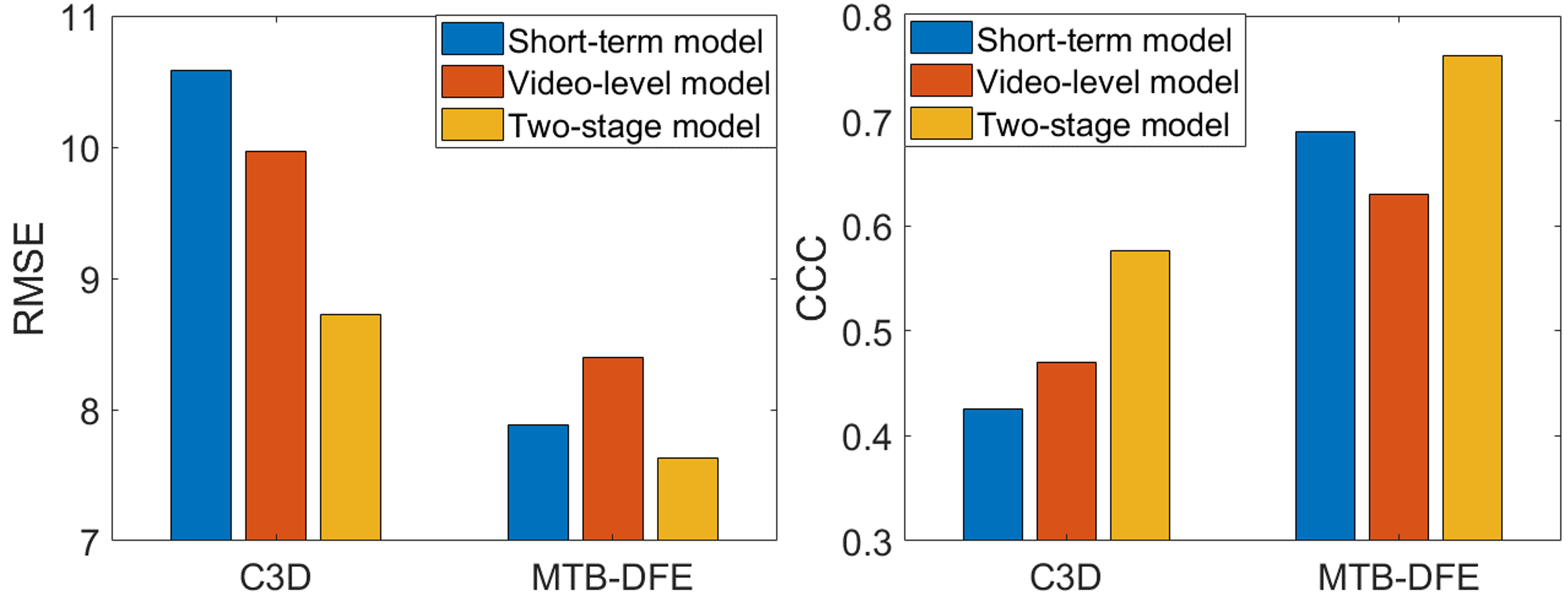}
\caption{Comparison of the results achieved by applying C3D and MTB-DFE for short-term depression modelling, long-term depression modelling, and two-stage depression modelling (combined with SPG) on AVEC 2014 Freeform dataset.}
\label{fig:result_c3d_mtb}
\end{figure}

Finally, we compare the results of C3D and MTB-DFE when using them for short-term modelling, video-level modelling and two-stage modelling. As illustrated in Fig. \ref{fig:result_c3d_mtb}, two-stage systems (the results achieved by applying C3D/MTB-DFE as the short-term model and then use SPG for video-level modelling) achieved the best results among all three settings for both networks, showing the clear advantages of the proposed two-stage framework. These results can be explained by the fact that the C3D/MTB-DFE-based video-level modelling discards short-term facial behaviour details during the down-sampling procedure, which may contain crucial cues for depression recognition. Meanwhile, when applying them for short-term depression modelling,   they fail to infer depression from video-level behaviours. In summary, we show that both short-term and video-level facial behaviour encoding are important for video-based depression recognition, suggesting the great potential of applying and extending the proposed two-stage framework for video-based automatic depression analysis applications.


\subsection{Cross-dataset evaluation}
\label{subsec:cross-dataset}

\noindent To further evaluate the generalization capability of the proposed approach, we also report the cross-datasets evaluation results in Table. \ref{tb:cross-dataset}. We observe that the models trained on AVEC 2013 dataset performed well on two AVEC 2014 tasks, especially the pre-trained MTB-DFE model achieved the PCC and RMSE of $0.732$ and $8.04$, respectively. In contrast, the MTB-DF models trained on short videos from AVEC 2014 are less robust. In particular, the models trained on the Freeform videos generated much better results than the models trained on the NorthWind videos. Since the AVEC 2013 tasks and Freeform tasks are unmediated and complex while NorthWind videos were recorded in strongly controlled conditions, (i.e., it only requires participants to read a pre-defined paragraph in German), the AVEC 2013 videos and Freeform videos (especially AVEC 2013 videos) contain richer facial behaviours than NorthWind videos. As a result, we hypothesize that the models can extract more depression-related cues from AVEC 2013 videos and Freeform videos. In other words, it shows that the models trained on tasks that elicit more natural behaviours and responses provide better generalisation capacity.

It also can be observed that most MTB-DFE models trained on AVEC 2013 and Freeform tasks outperformed their corresponding MTB-DFE+SPG models. This can be explained by the fact that the MTB-DFE only focuses on predicting depression from short-term facial behaviours and different tasks may still trigger some similar short-term facial behaviours. However, the SPG model attempts to learn video-level facial behaviours, which means they largely depend on the global contexts of the task. Consequently, the MTB-DFE+SPG models have worse generalization capability for cross-datasets evaluation.




\begin{table}[t]
\begin{center}
\normalsize
\scalebox{0.88}{
\begin{tabular}{| c | c  c | c  c|}
\hline
Method & Training set & Test set & PCC & RMSE\\
\hline
MTB-DFE  &  AVEC 2013 & NorthWind &0.732 & 8.04 \\
MTB-DFE  &  AVEC 2013 & Freeform  & 0.633&9.09 \\
MTB-DFE  &  NorthWind & AVEC 2013 & 0.639& 17.59 \\
MTB-DFE  &  NorthWind & Freeform  & 0.514& 17.46\\
MTB-DFE  &  Freeform & AVEC 2013  & 0.693&8.29 \\
MTB-DFE  &  Freeform & NorthWind &0.683 & 8.49 \\

\hline
MTB-DFE+SPG  &  AVEC 2013 & NorthWind &0.770 & 8.18 \\
MTB-DFE+SPG &  AVEC 2013 & Freeform  &0.689 & 8.62\\
MTB-DFE+SPG  &  NorthWind & AVEC 2013  &-0.238 & 16.56\\
MTB-DFE+SPG &  NorthWind & Freeform &-0.210 & 15.78\\
MTB-DFE+SPG  &  Freeform & AVEC 2013 & 0.650&9.13  \\
MTB-DFE+SPG &  Freeform & NorthWind &0.613 & 9.90 \\

\hline
\end{tabular}
}
\end{center}
\caption{Cross-dataset evaluation results.}
\label{tb:cross-dataset}
\end{table}

\subsection{Conclusions and discussion}

\noindent In this paper, we propose a specific, two-stage framework for video-based automatic depression recognition, where the first stage models depression from short-term facial behaviours and the second stage aims to construct a video-level depression representation based on all short-term facial behaviours of the target video, summarising long-term behavioural information. In particular, this paper proposes a MTB-DFE model to learn depression-related features from multi-scale short-term facial behaviours, which disentangles feature representations thereby enhancing the depression-related cues and removing non-depression noise encoded by the features. Here, we propose the first work to represent all short-term depression-related cues of the video as a graph representation for video-based depression analysis, i.e., SEG and SPG, both of which not only encode all thin slice-level features of the target video without discarding any frames, but also can be directly processed by GNNs for depression recognition. In other words, the proposed two-stage framework encodes depression cues from multi-scale short-term and long-term facial behaviours and provides the target depression prediction based on the behaviours portrayed by the entire video.

According to the experimental results on AVEC 2013 and AVEC 2014, we conclude that: (i). the proposed two-stage approach outperformed most existing methods with marginal advantages; (ii). the proposed MTB-DFE model also generated better performance than all existing short-term depression modelling methods, where the DFE module largely enhanced the performance, showing its capability to enhance depression-related cues and removing non-depression noises; (iii). both video-level graph representations can further improve the depression recognition performance, where SPGs produced better predictions than SEGs and other baselines, suggesting it may be a superior strategy for summarizing arbitrary number of thin slice-level features of a video; and (iv). the proposed two-stage framework can be easily extended using various short-term and long-term modelling methods. In particular, we found that under the same setting, two-stage modelling always provided better predictions than the corresponding one-stage methods.

While the proposed two-stage framework achieved the best and the most robust performance in depression recognition, a main limitation is that these two stages are implemented separately, which means the deep-learned short-term depression features may still not be optimal. If the short-term depression modelling and video-level depression encoding can be integrated into an end-to-end framework, both short-term and video-level depression representations could be potentially improved and produce better predictions. In addition, the existing audio-visual AVEC datasets only contain $150$ clips (the datasets used in \cite{ringeval2019avec} and \cite{ringeval2017avec} do not provide videos) and these datasets were collected in controlled lab environments. Thus, these experiments can not fully validate the usefulness of the proposed method for real-world applications. Consequently, a important future work in the field is to collect a larger real-world audio-visual dataset and provide it for public research usage by the community. Finally, since the DFE module and SPG achieved significant gains depression analysis, it would be interesting to extend them to similar video-level/clip-level recognition tasks, e.g., human action recognition and personality recognition.


%




\ifCLASSOPTIONcaptionsoff
  \newpage
\fi



%



\bibliographystyle{IEEEtran}

\bibliography{egbib}

%

\begin{IEEEbiography}[{\includegraphics[width=1in,height=1.25in,clip,keepaspectratio]{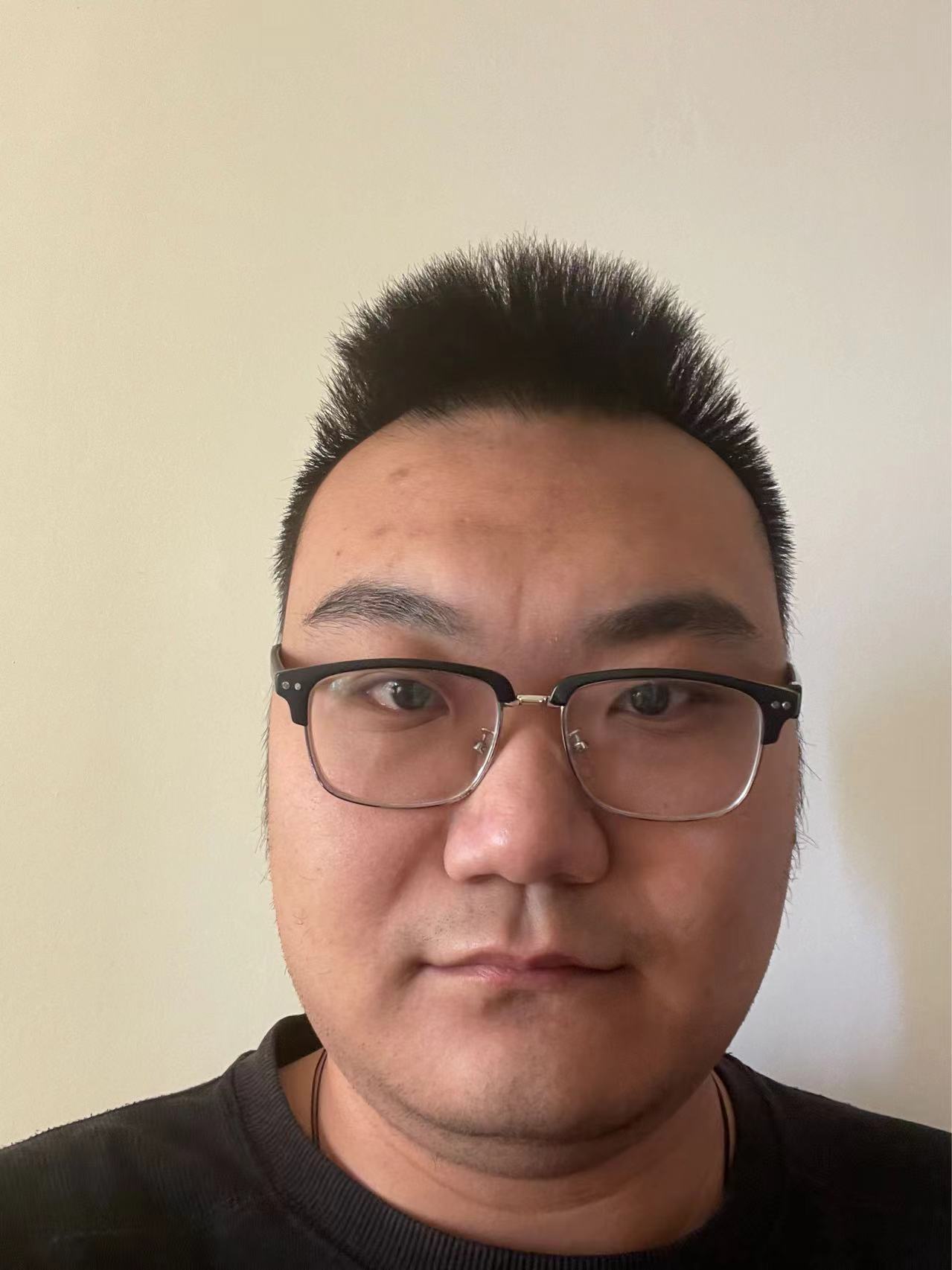}}]
{Jiaqi Xu} is a graduate student in the Sensors Information Lab at the Harbin University of Science and Technology, majoring in Pattern Recognition and Intelligent Systems. His research interests include depression analysis with various methods. 
\end{IEEEbiography}

\begin{IEEEbiography}
[{\includegraphics[width=1in,height=1.25in,clip,keepaspectratio]{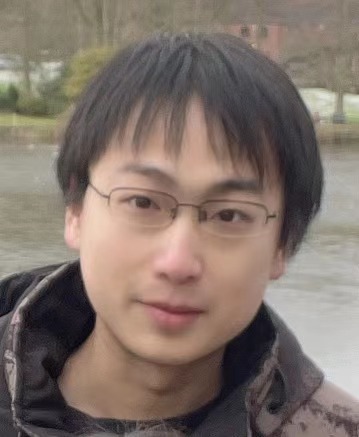}}]{Siyang Song} is a Postdoc researcher with the Department of Computer Science and Technology, University of Cambridge, Cambridge, U.K.. He received his PhD in the Computer Vision Lab and Horizon Center for Doctoral Training at the University of Nottingham, UK. His research interests include automatic emotion, personality and depression analysis by developing various self-supervised learning, Neural Architecture Search and graph modelling techniques. 
\end{IEEEbiography}

\begin{IEEEbiography}
[{\includegraphics[width=1in,height=1.25in,clip,keepaspectratio]{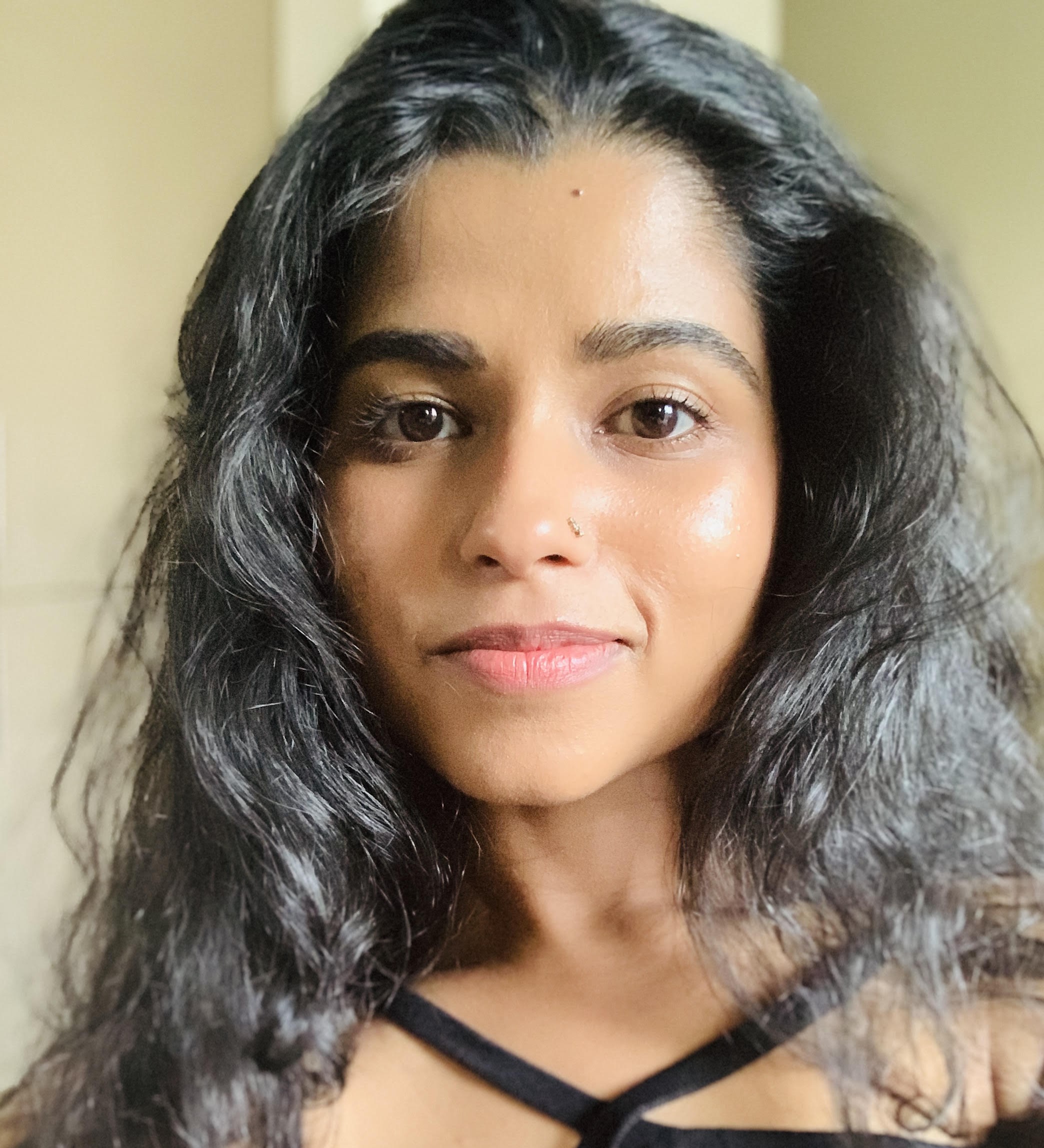}}]{Keerthy Kusumam}  is a PhD student at the Computer Vision Laboratory, Department of Computer Science, University of Nottingham, U.K., affiliated to the Horizon CDT. Her PhD focuses on the automatic assessment of mood disorders in videos collected under challenging and realistic scenarios. Her research experience includes topics in human pose estimation, face alignment, super-resolution, image synthesis using GANs, emotion and depression recognition. She attained her taught MSc in computer science from the University of Leeds and MSc by Research in computer vision and machine learning from the University of Lincoln.
\end{IEEEbiography}

\begin{IEEEbiography}[{\includegraphics[width=1in,height=1.25in,clip,keepaspectratio]{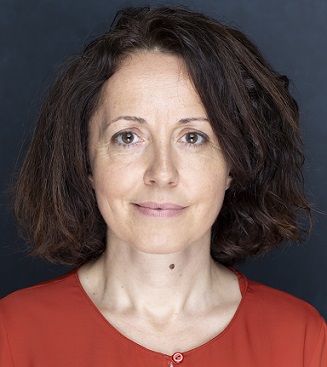}}]{Hatice Gunes} is a Professor with the Department of Computer Science and Technology, University of Cambridge, U.K., leading the Affective Intelligence and Robotics Lab. Her expertise is in the areas of affective computing and social signal processing cross-fertilizing research in human behavior understanding, computer vision, machine learning, and human–robot interaction. She has published over 125 papers in the above areas, and her research highlights  include  RSJ/KROS  Distinguished  Interdisciplinary Research Award Finalist at IEEE RO-MAN’21, Distinguished PC  Award  at  IJCAI’21,  Best  Paper  Award  Finalist  at  IEEE RO-MAN’20, Finalist for the 2018 Frontiers Spotlight Award, Outstanding  Paper  Award  at  IEEE  FG’11,  and  Best  Demo Award at IEEE ACII’09. Prof Gunes is the former President of the Association for the Advancement of Affective Computing (AAAC), and is a member of the Human-Robot Interaction Steering Committee. In 2019 she was awarded the prestigious EPSRC Fellowship as a personal grant and was named a Faculty Fellow of the Alan Turing Institute. 
\end{IEEEbiography}

\begin{IEEEbiography}[{\includegraphics[width=1in,height=1.25in,clip,keepaspectratio]{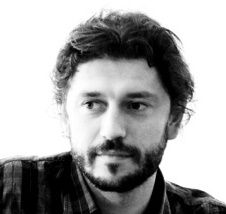}}]{Michel Valstar} is a Professor
in the Computer Vision and Mixed Reality Labs
at the University of Nottingham. He received his
masters degree in Electrical Engineering at Delft
University of Technology in 2005 and his PhD in
computer science with the intelligent Behaviour
Understanding Group (iBUG) at Imperial College
London in 2008. His main interest is in automatic
recognition of human behaviour. In 2011 he was
the main organiser of the first facial expression
recognition challenge, FERA 2011. In 2007 he
won the BCS British Machine Intelligence Prize for part of his PhD
work. He has published technical papers at authoritative conferences
including CVPR, ICCV and SMC-B and his work has received popular
press coverage in New Scientist and on BBC Radio.
\end{IEEEbiography}






\end{document}